\pdfoutput=1

\documentclass[11pt]{article}

\usepackage{acl}

\usepackage{times}
\usepackage{latexsym}
\usepackage[T1]{fontenc}\usepackage[utf8]{inputenc}
\usepackage{microtype}

\usepackage{amsmath,amsfonts,bm}

\def\eqref#1{equation~\ref{#1}}

\def\1{\bm{1}}

\DeclareMathAlphabet{\mathsfit}{\encodingdefault}{\sfdefault}{m}{sl}
\SetMathAlphabet{\mathsfit}{bold}{\encodingdefault}{\sfdefault}{bx}{n}

\usepackage{hyperref}
\usepackage{url}
\usepackage{booktabs}       
\usepackage{amsfonts}       
\usepackage{nicefrac}       
\usepackage{xcolor}         
\usepackage{enumitem}
\newcommand{\xhdr}[1]{{\noindent\bfseries #1}.}
\usepackage{subfigure}
\usepackage{graphicx}
\usepackage{multirow}
\usepackage{utfsym}
\usepackage{diagbox}
\usepackage{colortbl}
\usepackage[most]{tcolorbox}
\usepackage{caption}
\usepackage{wrapfig}
\definecolor{mypink}{rgb}{.99,.91,.95}
\definecolor{mygreen}{rgb}{.9,.99,.9}
\definecolor{mygray}{gray}{.9}
\usepackage{marvosym}
\usepackage{framed}
\usepackage{mdframed}
\usepackage{pifont}
\usepackage{CJKutf8}

\title{LongBench: A Bilingual, Multitask Benchmark \\for Long Context Understanding}

\author{Yushi Bai$^{12}$, Xin Lv$^{2}$, Jiajie Zhang$^{12}$, Hongchang Lyu$^3$, Jiankai Tang$^1$, \\ 
  \textbf{Zhidian Huang$^1$, Zhengxiao Du$^{12}$, Xiao Liu$^{12}$, Aohan Zeng$^{12}$,} \\ 
  \textbf{Lei Hou$^1$, Yuxiao Dong$^{1\dagger}$, Jie Tang$^1$, Juanzi Li$^{1\dagger}$} \\
  $^1$Tsinghua University
  \quad
  $^2$Zhipu.AI
  \quad
  $^3$Institute of Automation, Chinese Academy of Sciences}

\begin{document}
\maketitle

\renewcommand{\thefootnote}{\fnsymbol{footnote}}
    \footnotetext[2]{Corresponding authors
    }
\renewcommand{\thefootnote}{\arabic{footnote}}

\begin{abstract}

Although large language models (LLMs) demonstrate impressive performance for many language tasks, most of them can only handle texts a few thousand tokens long, limiting their applications on longer sequence inputs, such as books, reports, and codebases.
Recent works have proposed methods to improve LLMs' long context capabilities by extending context windows and more sophisticated memory mechanisms.
However, comprehensive benchmarks tailored for evaluating long context understanding are lacking.
In this paper, we introduce LongBench, the first bilingual, multi-task benchmark for long context understanding, enabling a more rigorous evaluation of long context understanding.
LongBench comprises 21 datasets across 6 task categories in both English and Chinese, with an average length of 6,711 words (English) and 13,386 characters (Chinese). 
These tasks cover key long-text application areas including single-doc QA, multi-doc QA, summarization, few-shot learning, synthetic tasks, and code completion. 
All datasets in LongBench are standardized into a unified format, allowing for effortless automatic evaluation of LLMs.
Upon comprehensive evaluation of 8 LLMs on LongBench, we find that: (1) Commercial model (GPT-3.5-Turbo-16k) outperforms other open-sourced models, but still struggles on longer contexts. (2) Scaled position embedding and fine-tuning on longer sequences lead to substantial improvement on long context understanding. (3) Context compression technique such as retrieval brings improvement for model with weak ability on long contexts, but the performance still lags behind models that have strong long context understanding capability.

\end{abstract}

\section{Introduction}
\label{sec:intro}

\begin{figure*}
    \centering
    \includegraphics[width=0.9\linewidth]{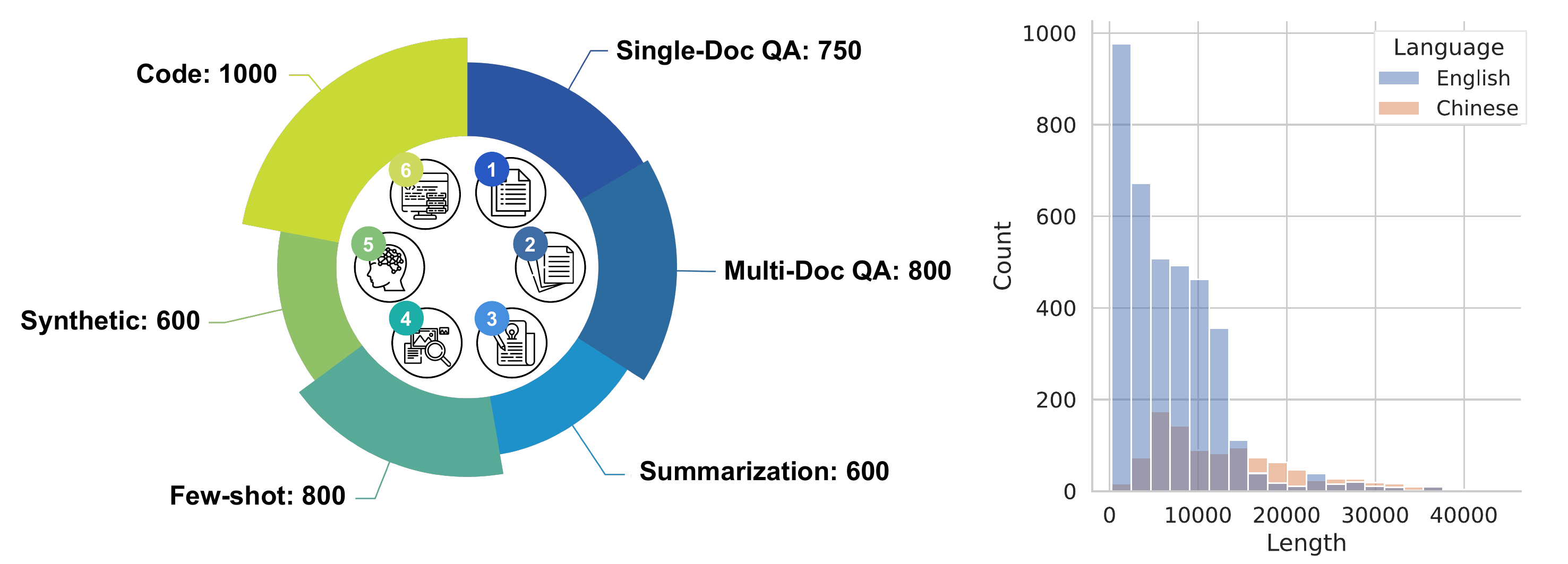}
    \caption{Left: Number of data in each type of task within LongBench. Right: Length distribution for English and Chinese data in LongBench, measured by the number of words and characters.}
    \label{fig:overview}
\end{figure*}

The field of NLP has long sought to endow machines with the ability to understand and reason over a long context.
Tasks such as summarization and question answering based on books, reports, and documents, and code generation at the repository level demand the ability to model long context sequences that span thousands or even tens of thousands of tokens in length. However, many of today's large language models can only comprehend and generate texts a few thousand tokens long, leaving room for potential improvements in processing longer contexts.
More recently, there has been an increasing effort to improve large language models' capabilities on long context understanding.
These methods include extending the context window~\citep{press2022train,chen2023extending}, utilizing recurrent memory~\citep{dai2019transformer,bulatov2023scaling}, using sparsed attention~\citep{ding2023longnet,mohtashami2023landmark}, and augmenting with an external memory~\citep{liang2023unleashing,zhou2023recurrentgpt}.
However, unlike in short context, where a multitude of multi-task benchmarks are available for a multi-aspect evaluation~\citep{hendrycks2021measuring,srivastava2023beyond}, there is no such benchmark on longer context.

To facilitate further research in this direction, we propose \textbf{LongBench}, the first bilingual, multi-task benchmark tailored for long context understanding.
LongBench is composed of 6 major task categories and 21 different tasks, covering key long-text application scenarios including multi-document QA, single-document QA, summarization, few-shot learning, code completion, and synthetic tasks.
In addition, LongBench includes different languages (Chinese and English) to provide a more comprehensive evaluation of the large models' bilingual capabilities on long contexts.
With overview statistics shown in Figure~\ref{fig:overview}, LongBench contains 4,750 test instances, with an average length of 6,711 words and 13,386 characters for English and Chinese instances, respectively.

All 21 datasets in LongBench are standardized into a unified format, among which 6 are directly extracted from the original datasets provided by previous studies, 10 are built based on the original datasets and processed to be suitable for long context evaluation, and 5 are created and annotated by us.
We are fully aware of the potentially high costs involved in the model evaluation process, especially in the context of long context scenarios (such as manual annotation costs or API call costs). Therefore, we adopt a fully automated evaluation method, where we utilize automatic metrics such as ROUGE-L and F1 to measure the similarity of the outputs to the groundtruth answers.

We conduct a comprehensive evaluation of 8 models on LongBench. The empirical results provide insightful conclusions about the multi-task capability of current models in terms of long context comprehension.
Additionally, to better disentangle the models' long context ability from their task ability, we construct LongBench-E that features a more \textit{even} length distribution, thus suited for gauging each model's capability across various context lengths.
Results on LongBench-E reveal that although some models are trained or fine-tuned on longer contexts, they still experience a significant decline in performance as the context length increases.
We also investigate the effect of retrieval-based and summarization-based context compression techniques. Our results demonstrate that these approaches are beneficial only to models that exhibit weaker capability on long contexts.

\section{Related Work}
\label{sec:related}

\xhdr{Long Context Modeling Techniques}
We first discuss some popular lines of methods that aim to tackle long context understanding.
These studies are mainly aimed at solving two key challenges in long text modeling, including the high runtime overhead on longer context, and the catastrophic forgetting phenomenon when processing long sequence.
A series of studies focus on how to make Transformers more efficient and unforgetful~\citep{tay2022efficient}, with designs such as sparse and efficient computation~\citep{child2019generating,kitaev2020reformer,beltagy2020longformer,zaheer2020big,wang2020linformer,fedus2022switch,ding2023longnet}, recurrent and memory modules~\citep{dai2019transformer,rae2020compressive,wu2022memorizing,martins2022former,bulatov2022recurrent,orvieto2023resurrecting,liang2023unleashing,zhou2023recurrentgpt}.
More recently, several methods~\citep{press2022train,sun2022length,chen2023extending} have been proposed to enable length extrapolation of Transformers, and have been adopted in the training process of long context LLMs such as ChatGLM2-32k~\citep{zeng2022glm} and LongChat-32k~\citep{longchat2023}.

\xhdr{Evaluation for Long Context Understanding}
Many previous works on long text modeling rely on the perplexity metric for evaluation~\citep{beltagy2020longformer,roy2021efficient,press2022train}.
However, as suggested in \cite{sun2021long}, the perplexity metric may not necessarily reflect the model's performance on sequence-level tasks in real applications
Meanwhile, some works assess long text modeling through artificial tasks such as retrieval~\citep{tay2021long,chen2023extending,longchat2023}, which may also fall short in mirroring real-world scenarios.

Concurrently, ZeroSCROLLS~\citep{shaham2022scrolls,shaham2023zeroscrolls} and L-Eval~\citep{an2023eval} are proposed as evaluation benchmarks for long text modeling.
Nonetheless, they encompass a restricted range of task types, thereby limiting the diversity of long text modeling patterns required in the benchmarks, and consequently, the comprehensiveness of the evaluation results.
Recently, AgentBench~\citep{liu2023agentbench} also mentions the challenge of LLM-as-Agent's handling long interaction trajectories, but fails to incorporate it as a dedicated evaluation dimension.
In contrast, LongBench includes six major task categories, with each category featuring sequences of varying lengths, languages, and domains. We believe it provides a more holistic evaluation of long text modeling ability of large language models across a spectrum of lengths, distributions, as well as long dependency patterns.

\section{LongBench: Task and Construction}
\label{sec:method}

\begin{table*}[t]
\centering  
\resizebox{0.95\textwidth}{!}{
\begin{tabular}{lclrccc}
\toprule
Dataset & ID & Source & Avg len & Metric & Language & \#data \\
\midrule
\emph{Single-Document QA} \\
NarrativeQA & 1-1 & Literature, Film & 18,409 & F1 & English & 200 \\
Qasper & 1-2 & Science & 3,619 & F1 & English & 200 \\
MultiFieldQA-en & 1-3 & Multi-field & 4,559 & F1 & English & 150 \\
\cellcolor{mypink}MultiFieldQA-zh & \cellcolor{mypink}1-4 & \cellcolor{mypink}Multi-field & \cellcolor{mypink}6,701 & \cellcolor{mypink}F1 & \cellcolor{mypink}Chinese & \cellcolor{mypink}200 \\
\midrule
\emph{Multi-Document QA} \\
HotpotQA & 2-1 & Wikipedia & 9,151 & F1 & English & 200 \\
2WikiMultihopQA & 2-2 & Wikipedia & 4,887 & F1 & English & 200 \\
MuSiQue & 2-3 & Wikipedia & 11,214 & F1 & English & 200 \\
\cellcolor{mypink}DuReader & \cellcolor{mypink}2-4 & \cellcolor{mypink}Baidu Search & \cellcolor{mypink}15,768 & \cellcolor{mypink}Rouge-L & \cellcolor{mypink}Chinese & \cellcolor{mypink}200 \\
\midrule
\emph{Summarization} \\
GovReport & 3-1 & Government report & 8,734 & Rouge-L & English & 200 \\
QMSum & 3-2 & Meeting & 10,614 & Rouge-L & English & 200 \\
MultiNews & 3-3 & News & 2,113 & Rouge-L & English & 200 \\
\cellcolor{mypink}VCSUM & \cellcolor{mypink}3-4 & \cellcolor{mypink}Meeting & \cellcolor{mypink}15,380 & \cellcolor{mypink}Rouge-L & \cellcolor{mypink}Chinese & \cellcolor{mypink}200 \\
\midrule
\emph{Few-shot Learning} \\
TREC & 4-1 & Web question & 5,177 & Accuracy (CLS) & English & 200 \\
TriviaQA & 4-2 & Wikipedia, Web & 8,209 & F1 & English & 200 \\
SAMSum & 4-3 & Dialogue & 6,258 & Rouge-L & English & 200 \\
\cellcolor{mypink}LSHT & \cellcolor{mypink}4-4 & \cellcolor{mypink}News & \cellcolor{mypink}22,337 & \cellcolor{mypink}Accuracy (CLS) & \cellcolor{mypink}Chinese & \cellcolor{mypink}200 \\
\midrule
\emph{Synthetic Task} \\
PassageCount & 5-1 & Wikipedia & 11,141 & Accuracy (EM) & English & 200 \\
PassageRetrieval-en & 5-2 & Wikipedia & 9,289 & Accuracy (EM) & English & 200 \\
\cellcolor{mypink}PassageRetrieval-zh & \cellcolor{mypink}5-3 & \cellcolor{mypink}C4 Dataset & \cellcolor{mypink}6,745 & \cellcolor{mypink}Accuracy (EM) & \cellcolor{mypink}Chinese & \cellcolor{mypink}200 \\
\midrule
\emph{Code Completion} \\
LCC & 6-1 & Github & 1,235 & Edit Sim & Python/C\#/Java & 500 \\
RepoBench-P & 6-2 & Github repository & 4,206 & Edit Sim & Python/Java & 500 \\
\bottomrule
\end{tabular}
}
\caption{An overview of the dataset statistics in LongBench. Chinese datasets are highlighted. `Source' denotes the origin of the context. `Avg len' (average length) is computed using the number of words for the English (code) datasets and the number of characters for the Chinese datasets. `Accuracy (CLS)' refers to classification accuracy, while `Accuracy (EM)' refers to exact match accuracy.}
\label{tb:stat}
\vspace{-5mm}
\end{table*}

\subsection{Problem Definition}
We formalize the problem of long context understanding as follows: Given the input and context sequences $(I, C)$, the model is expected to output the answer $A$.
For instance, in a QA task, the input $I$ would be the question, context $C$ refers to the document, and $A$ denotes the answer to the question.
Generally, in LongBench, $I$ and $A$ tend to be short, while $C$ represents a long sequence up to thousands of tokens in length. The instantiation of $(I, C, A)$ for each task is listed in Table~\ref{tb:form}.

\subsection{Dataset Construction}
In this section, we will provide a detailed introduction to the data collection, annotation, and organization process for each dataset in LongBench (LongBench-E), according to the specific tasks.
For the overall data statistics of LongBench, we refer to Table~\ref{tb:stat}.

\subsubsection{Data Collection and Annotation}

\xhdr{Single-Doc QA}
For single-document QA, we focus on instances with longer documents.
We extract \emph{NarrativeQA} from the original dataset in \citet{kovcisky2018narrativeqa}, which consists of long stories along with questions posed to test reading comprehension.
We also sample from \emph{Qasper}~\citep{dasigi2021dataset}, which features QA over NLP papers and is annotated by NLP practitioners.

To better test the model's long context understanding ability across diverse fields, we manually curate the \emph{MultiFieldQA} datasets in both English and Chinese.
We first collect documents and articles from multiple sources, including legal documents, government reports, encyclopedias, academic papers, etc (sources detailed in Appendix). We invite three Ph.D. students to annotate the question and answer for each article, with definitive answers as much as possible for ease of automated evaluation. During the annotation, we ensure that the answers can be inferred from the documents, and the placement of evidence is fairly random to avoid biases that might occur if, for instance, the answer-related statements are frequently found at the beginning or the end, as mentioned in \cite{liu2023lost}.

\xhdr{Multi-Doc QA}
Multi-document QA requires models to extract and combine information from several documents to obtain the answer, which is usually more challenging than single-doc QA. The English test samples are built from three Wikipedia-based multi-hop QA datasets: \emph{HotpotQA}~\citep{yang2018hotpotqa}, \emph{2WikiMultihopQA}~\citep{ho2020constructing}, and \emph{MuSiQue}~\citep{trivedi2022musique}. HotpotQA involves a number of 2-hop questions directly written by native speakers given two related paragraphs. 2WikiMultihopQA consists of up to 5-hop questions that are synthesized through manually designed templates to ensure that they cannot be solved through shortcuts.
The questions in MuSiQue are carefully composed from simple questions involving up to 4-hop reasoning, and are then paraphrased by annotators to both avoid shortcuts and ensure linguistic naturalness.
Each question in the original datasets is supplemented by 2-4 supporting paragraphs that provide one-step reasoning evidence and several distracting paragraphs.

To tailor the data for long-context evaluation, we utilize complete Wikipedia passages encompassing the supporting or distracting paragraphs as the context. Initially, supporting passages are included within the context, and then as many distracting passages are added until the total length reaches a maximum length. Finally, these passages are randomly ordered to form the multi-document context.

Beyond these three English datasets, we also construct a Chinese dataset based on \emph{DuReader}~\citep{he2018dureader}, which is developed based on Baidu Search and Baidu Zhidao, comprising 200K questions and 1M related documents. To adapt it for assessing long context ability, for each question, we not only provide several documents related to the question but also arbitrarily select several from the total set of documents as distractors, until each question is associated with 20 documents.

\xhdr{Summarization}
Compared to QA tasks, which can often be solved using local information within the context, summarization demands a more global understanding of the whole context.
We extract \emph{GovReport} from the original dataset~\citep{huang2021efficient}.
The original GovReport dataset is a large-scale collection of detailed reports from the U.S. Government Accountability Office and Congressional Research Service, each accompanied by a human-written summary, spanning a wide variety of national policy issues.
We also sampled from \emph{QMSum}~\cite{zhong2021qmsum}, which consists of query-summary pairs annotated over 232 meetings across multiple domains, including product, academic, and committee meetings.
We treat the query as input $I$, the meeting content as context $C$, and the summary as answer $A$.
\emph{MultiNews} is derived from the original multi-document summarization dataset in \cite{fabbri2019multi}.
The MultiNews dataset features clusters of 2-10 news articles discussing the same event or topic, each paired with a human-written summary that summarizes the key information from the multiple source articles.
In LongBench, we include ``Document $i$'' before the $i$th news article, and concatenate them into the context $C$.
\emph{VCSUM}~\citep{wu-etal-2023-vcsum} is a large-scale Chinese meeting summarization dataset consisting of 239 real-life meetings with over 230 hours of duration, with versatile annotations to support multiple summarization tasks. In LongBench, we select the long segments from VCSUM to compose our evaluation samples.

\xhdr{Few-shot Learning}
We identify few-shot in-context learning as a practical setting that requires long context understanding, especially when the number of examples increases~\citep{ainslie2023colt5}.
To ensure the diversity of the tasks, we incorporate classification, summarization, and reading comprehension tasks within the few-shot learning scenario.
We include 2 classification datasets with fine-grained class labels, including \emph{TREC}~\citep{li2002learning}, a question classification task involving 50 fine classes, and \emph{LSHT}~\citep{lsht}, a Chinese news classification task with 24 classes. 
For summarization task, we use the \emph{SAMSum} dataset~\citep{gliwa2019samsum}, which contains messenger-like conversations with human-annotated summaries.
\emph{TriviaQA}~\citep{joshi2017triviaqa} contains question-answer pairs labeled with evidence passages, and we use it as a reading comprehension task.
We filter the passages in TriviaQA with less than 1,000 words to be examples.

In each of the above datasets adapted for LongBench, for each test data, we first randomly select an integer within a range as the number of examples, then randomly sample the corresponding number of samples from the training set, and concatenate them to form the context $C$. For TREC, LSHT, SAMSum, and TriviaQA, the ranges are $[100, 600], [10, 40], [10, 100], [2, 24]$, respectively.

\xhdr{Synthetic Task}
Unlike standard tasks that are more alike on the required long dependency pattern, synthetic tasks can be meticulously designed to test the model's ability on specific scenarios and patterns. In LongBench, we design three synthetic tasks.
\emph{PassageRetrieval-en} and \emph{PassageRetrieval-zh} are constructed based on English Wikipedia and the Chinese sections of the C4 dataset~\citep{t5}. For each data entry, we randomly sample 30 passages and select one of them for summarization using GPT-3.5-Turbo. The task asks the model to identify the original paragraph to which the crafted summary corresponds.

\emph{PassageCount} seeks to create a more demanding situation where the model is required to utilize the full context to resolve the task.
For each piece of data, we randomly select several passages from English Wikipedia, repeat each paragraph at random several times, and finally shuffle the paragraphs. The task asks the model to determine the number of unique passages among the given set. Specifically, we randomly select $M$ from $[17, 50]$ as the upper limit for the number of passages. Subsequently, the number $N$ of unique passages is randomly selected from the range $[2, M]$. We conduct random sampling with replacement from the set of $N$ unique passages to get the final $M$ passages.

\xhdr{Code Completion}
Code completion is a critical task employed by auto-completion systems to assist users by completing code based on previous code input and context~\citep{chen2021evaluating,zheng2023codegeex}.
This task can pose a significant challenge for models, especially when dealing with lengthy code inputs or even repository-level data. 
This is mainly because the models need to establish attention across long-range sequences according to relationships within code elements, such as between class and function definitions.
Hence we recognize this as a suitable task for evaluating a model's long context modeling ability.

The \emph{LCC} dataset is sampled from the original Long Code Completion dataset~\citep{guo2023longcoder}. The original dataset is constructed by filtering code within one file from GitHub based on length.
This data includes a long piece of preceding lines of code as context, and the next line of code as the answer.
We also consider the repository-level code completion setting, which necessitates aggregating information from code across files.
For this task, we adapt the \emph{RepoBench-P} dataset from \cite{liu2023repobench}.
RepoBench-P is collected from Github repositories, and is constructed by first retrieving relevant code snippets from other files based on module import statements. These snippets are then concatenated with the preceding lines of code within the current file as context, and are used to predict the next line of code.
We select the most challenging XF-F (Cross-File-First) setting from the original dataset, where the in-file context gives no prior usage of the module to aid the prediction.
For each original piece of data, we shuffle the cross-file code snippets that include the gold cross-file code snippet (manually annotated as the optimal context for prediction), and combine them into context $C$.
The preceding lines of code are taken as input $I$, and the next line of code as the answer $A$.

\begin{table*}[t]
\centering  
\resizebox{\textwidth}{!}{
\begin{tabular}{l|ccccc|ccccc|ccccc}
\toprule
\multirow{2}{*}{\textbf{Model}} & \multicolumn{5}{c|}{\textbf{Single-Doc QA}} & \multicolumn{5}{c|}{\textbf{Multi-Doc QA}} & \multicolumn{5}{c}{\textbf{Summarization}} \\
\cmidrule(lr){2-6} \cmidrule(lr){7-11} \cmidrule(lr){12-16} 
& \textbf{1-1} & \textbf{1-2} & \textbf{1-3} & \textbf{1-4} & \textbf{Avg} & \textbf{2-1} & \textbf{2-2} & \textbf{2-3} & \textbf{2-4} & \textbf{Avg} & \textbf{3-1} & \textbf{3-2} & \textbf{3-3} & \textbf{3-4} & \textbf{Avg} \\
\midrule
GPT-3.5-Turbo-16k & 23.6 & 43.3 & 52.3 & 61.2 & 45.1 & 51.6 & 37.7 & 26.9 & 28.7 & 36.2 & 29.5 & 23.4 & 26.7 & 16.0 & 23.9 \\
\rowcolor{mygray}Llama2-7B-chat-4k & 18.7 & 19.2 & 36.8 & 11.9 & 21.7 & 25.4 & 32.8 & 9.4 & 5.2 & 18.2 & 27.3 & 20.8 & 25.8 & 0.2 & 18.5 \\
LongChat-v1.5-7B-32k & 16.9 & 27.7 & 41.4 & 29.1 & 28.8 & 31.5 & 20.6 & 9.7 & 19.5 & 20.3 & 30.8 & 22.7 & 26.4 & 9.9 & 22.5 \\
\rowcolor{mygray}XGen-7B-8k & 18.0 & 18.1 & 37.7 & 14.8 & 22.1 & 29.7 & 21.1 & 10.3 & 11.0 & 18.0 & 27.3 & 20.5 & 26.2 & 2.2 & 19.0 \\
InternLM-7B-8k & 12.1 & 16.7 & 23.4 & 33.6 & 21.4 & 28.7 & 22.8 & 9.0 & 11.1 & 17.9 & 9.7 & 15.9 & 22.8 & 12.4 & 15.2 \\
\rowcolor{mygray}ChatGLM2-6B & 11.8 & 22.5 & 35.0 & 33.2 & 25.6 & 22.4 & 20.1 & 6.1 & 16.3 & 16.2 & 23.2 & 21.1 & 25.2 & 14.5 & 21.0 \\
ChatGLM2-6B-32k & 21.1 & 31.5 & 46.2 & 51.6 & 37.6 & 45.1 & 34.0 & 21.9 & 37.6 & 34.7 & 32.4 & 24.0 & 26.5 & 16.2 & 24.8 \\
\rowcolor{mygray}Vicuna-v1.5-7B-16k & 19.4 & 26.1 & 38.5 & 43.0 & 31.8 & 25.3 & 20.8 & 9.8 & 19.3 & 18.8 & 27.9 & 22.8 & 27.2 & 15.1 & 23.2 \\
\bottomrule
\end{tabular}
}
\caption{Results (\%) on single-doc QA, multi-doc QA and summarization tasks.}
\label{tb:exp1}
\end{table*}


\begin{table*}[t]
\centering
\resizebox{\textwidth}{!}{
\begin{tabular}{l|ccccc|cccc|ccc|ccc}
\toprule
\multirow{2}{*}{\textbf{Model}} & \multicolumn{5}{c|}{\textbf{Few-shot Learning}} & \multicolumn{4}{c|}{\textbf{Synthetic}} & \multicolumn{3}{c|}{\textbf{Code}} & \multicolumn{3}{c}{\textbf{Overall}} \\
\cmidrule(lr){2-6} \cmidrule(lr){7-10} \cmidrule(lr){11-13} \cmidrule(lr){14-16}
& \textbf{4-1} & \textbf{4-2} & \textbf{4-3} & \textbf{4-4} & \textbf{Avg} & \textbf{5-1} & \textbf{5-2} & \textbf{5-3} & \textbf{Avg} & \textbf{6-1} & \textbf{6-2} & \textbf{Avg} & \textbf{EN} & \textbf{ZH} & \textbf{All} \\
\midrule
GPT-3.5-Turbo-16k & 68.0 & 91.4 & 41.7 & 29.2 & 57.6 & 4.5 & 71.0 & 77.5 & 51.0 & 54.7 & 53.6 & 54.1 & 44.0 & 44.5 & 44.7 \\
\rowcolor{mygray}Llama2-7B-chat-4k & 61.5 & 77.8 & 40.7 & 19.8 & 49.9 & 2.1 & 9.8 & 0.5 & 4.1 & 52.4 & 43.8 & 48.1 & 31.0 & 14.3 & 26.8 \\
LongChat-v1.5-7B-32k & 63.5 & 82.3 & 34.2 & 23.2 & 50.8 & 1.0 & 30.5 & 7.6 & 13.0 & 53.0 & 55.3 & 54.1 & 34.3 & 23.9 & 31.6 \\
\rowcolor{mygray}XGen-7B-8k & 65.5 & 77.8 & 25.3 & 20.5 & 47.3 & 2.1 & 8.5 & 3.5 & 4.7 & 38.6 & 38.6 & 38.6 & 28.3 & 15.1 & 25.0 \\
InternLM-7B-8k & 52.0 & 77.8 & 21.2 & 15.2 & 41.6 & 3.0 & 6.0 & 0.9 & 3.3 & 44.1 & 28.8 & 36.4 & 24.2 & 18.3 & 22.6 \\
\rowcolor{mygray}ChatGLM2-6B & 44.5 & 70.6 & 29.5 & 20.8 & 41.3 & 2.5 & 3.0 & 6.5 & 4.0 & 49.0 & 43.2 & 46.1 & 26.6 & 22.9 & 25.7 \\
ChatGLM2-6B-32k & 62.5 & 78.7 & 36.3 & 27.7 & 51.3 & 1.5 & 77.0 & 64.5 & 47.7 & 55.6 & 49.9 & 52.7 & 40.9 & 41.7 & 41.4 \\
\rowcolor{mygray}Vicuna-v1.5-7B-16k & 71.5 & 86.2 & 40.8 & 28.8 & 56.8 & 6.5 & 4.5 & 5.0 & 5.3 & 51.0 & 43.5 & 47.3 & 31.9 & 26.4 & 30.5 \\
\bottomrule
\end{tabular}
}
\caption{Results (\%) on few-shot learning, synthetic, and code tasks. `Overall' is computed by the macro-average (the mean of `Avg') over major task categories. This is computed on English (EN) tasks, Chinese (ZH) tasks, and all (All) tasks, code tasks are included in both languages.}
\label{tb:exp2}
\end{table*}

\subsubsection{Data Extraction}
\label{sec:ext}
Since LLMs may have already been trained on the training set of some of our collected public datasets, to avoid test leakage, we extract data from the test sets of these public datasets, with the exception of VCSUM due to its insufficient data in its test set.
We employ two extraction strategies: random sampling and uniform sampling. Through random sampling, we maintain a natural length distribution to more accurately mimic real scenarios, and obtain LongBench.
Alternatively, we perform uniform sampling based on the length of the data with a focus on studying the model's capabilities across varying context lengths within each task itself. This approach provides insights into the model's true ability to understand long contexts independent of task capability.
We choose 13 of the English datasets, including Qasper, MultiFieldQA-en, HotpotQA, 2WikiMultihopQA, GovReport, Multi-news, TREC, TriviaQA, SAMSum, PassageCount, PassageRetrieval-en, LCC, and RepoBench-P, which offer broader coverage on data length.
During the uniform sampling process, we use word count as the length and sample a comparable quantity of data from the length ranges of 0-4k, 4k-8k, and 8k+.
The resulting data is compiled into LongBench-E (statistics in Table~\ref{tb:e}).

\section{Experiments}
\label{sec:experiments}

\subsection{Benchmarking Results on LongBench and LongBench-E}

\xhdr{Experiment Setup}
We evaluate 8 popular LLMs that feature long context capability, including GPT-3.5-Turbo-16k~\citep{chatgpt}, Llama2-7B-chat-4k~\citep{touvron2023llama}, LongChat-v1.5-7B-32k~\citep{longchat2023}, XGen-7B-8k~\citep{XGen}, InternLM-7B-8k~\citep{2023internlm}, ChatGLM2-6B, ChatGLM2-6B-32k~\citep{du2022glm,zeng2022glm}, and Vicuna-v1.5-7B-16k~\citep{zheng2023judging}.
ChatGLM2-6B-32k is trained based on ChatGLM2-6B, with a 32k context length during alignment and position interpolation~\citep{chen2023extending}.
LongChat-v1.5-7B-32k and Vicuna-v1.5-7B-16k are fine-tuned from Llama2-7B, with supervised fine-tuning and linear RoPE scaling.

We conduct the assessment in a zero-shot setting, except for the few-shot learning tasks where the few-shot examples are provided as part of the long context.
The input format prompt and the maximum output length we used during evaluation can be found in Appendix.
When the input length $L$ surpasses the maximum context length $M$ of a model (indicated by the suffix of its name), we truncate the input sequence $S$ from the middle since the front and end of the sequence may contain crucial information such as the instruction or question:
$S_{1:L}\rightarrow [S_{1:\lfloor M/2\rfloor}; S_{L-\lfloor M/2\rfloor-1:L}]$.
During generation, we use greedy decoding for reproducibility.
It's worth noting that the chat models evaluated typically have specific prompts that induce the models to generate dialogue-like responses.
During evaluation, we avoid adding these prompts in few-shot learning and code completion tasks, since the answers to these tasks should be generated in a completion style rather than a chat style.

The metric for each dataset is shown in Table~\ref{tb:stat}. 
For tasks built based on previous datasets, the metrics we used are consistent with those used in the original work.
F1 and ROUGE-L~\citep{lin2004rouge} are two popular N-gram based metrics widely adopted in QA and summarization tasks.
Edit Sim (Levenshtein distance) is popularly used in code generation evaluation~\citep{svyatkovskiy2020intellicode}.
For the few-shot learning tasks, we extract the first line of the response.
For the two code completion tasks, we extract the first line of model generation that is not comment.
The code and datasets are available at \url{https://github.com/THUDM/LongBench}.

\xhdr{Results on LongBench}
Table~\ref{tb:exp1}, \ref{tb:exp2} report the performance (\%) on all datasets in LongBench. Additionally, Figure~\ref{fig:radar} presents a radar plot depicting the models' abilities on the 6 major tasks.
For better visualization, we scale the maximum score across all models on each task to 100 in the radar plot.
We summarize our key findings from the experiment results:
(1) There is still a performance gap on long context tasks between open-sourced models of smaller size and commercial model (GPT-3.5-Turbo-16k).
(2) Models benefit from scaled positional embedding and continued training on longer context, as ChatGLM2-6B-32k and LongChat-v1.5-7B-32k obtain relative improvements of 62\% and 19\%, respectively.
We further analyze the multi-task property of LongBench by the inter-task correlation among and across each category of tasks in Appendix~\ref{app:corr}. We find higher correlations for performance on tasks of the same category or language.

\begin{figure}[t]
    \centering
    \includegraphics[width=0.4\textwidth]{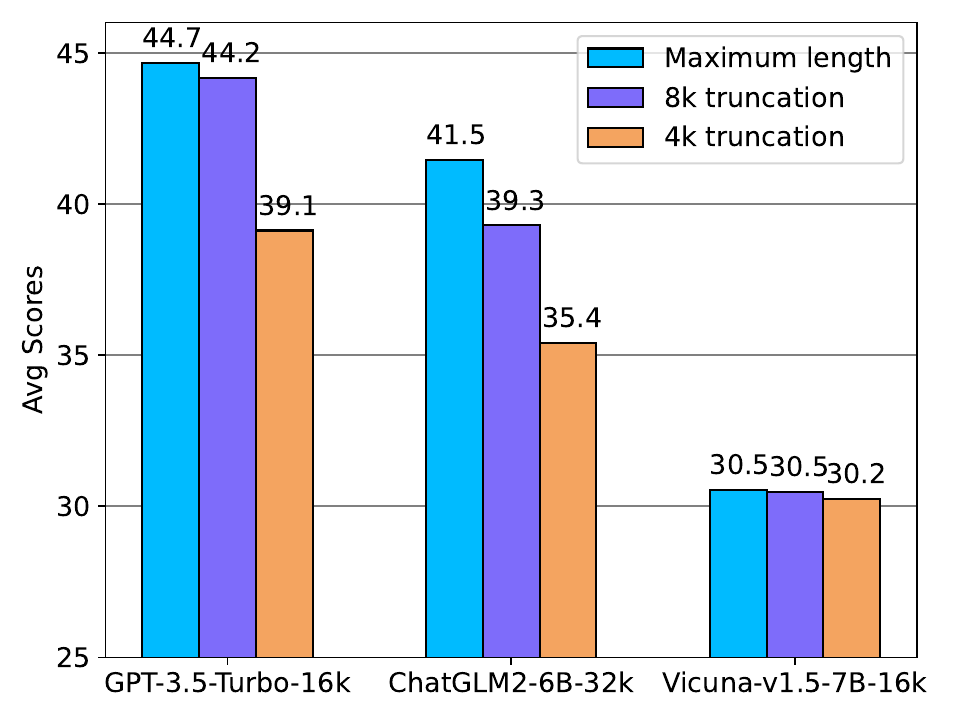}
    \caption{Avg score (\%) under different truncation size.}
    \label{fig:bar}
\end{figure}

To study whether the models with a longer maximum length truly benefit from utilizing longer context, we conduct experiments with GPT-Turbo-3.5-16k, ChatGLM2-6B-32k, and Vicuna-v1.5-7B-16k with truncation sizes of 4k and 8k on LongBench. The macro-average scores across all tasks with varying truncation sizes are depicted in Figure~\ref{fig:bar}. Here, `maximum length' denotes truncation at the model's maximum length configuration.
We observe that GPT-Turbo-3.5-16k and ChatGLM2-6B-32k obtain higher scores under a larger truncation size, suggesting they can better make use of a longer context.
Furthermore, this confirms that our benchmark indeed necessitates long context modeling --- using truncated information alone is insufficient for successfully completing the tasks in LongBench. On the other hand, the performance of LLMs on LongBench can be further improved by enhancing their long context modeling capabilities.

\begin{figure}[t]
    \centering
    \includegraphics[width=0.4\textwidth]{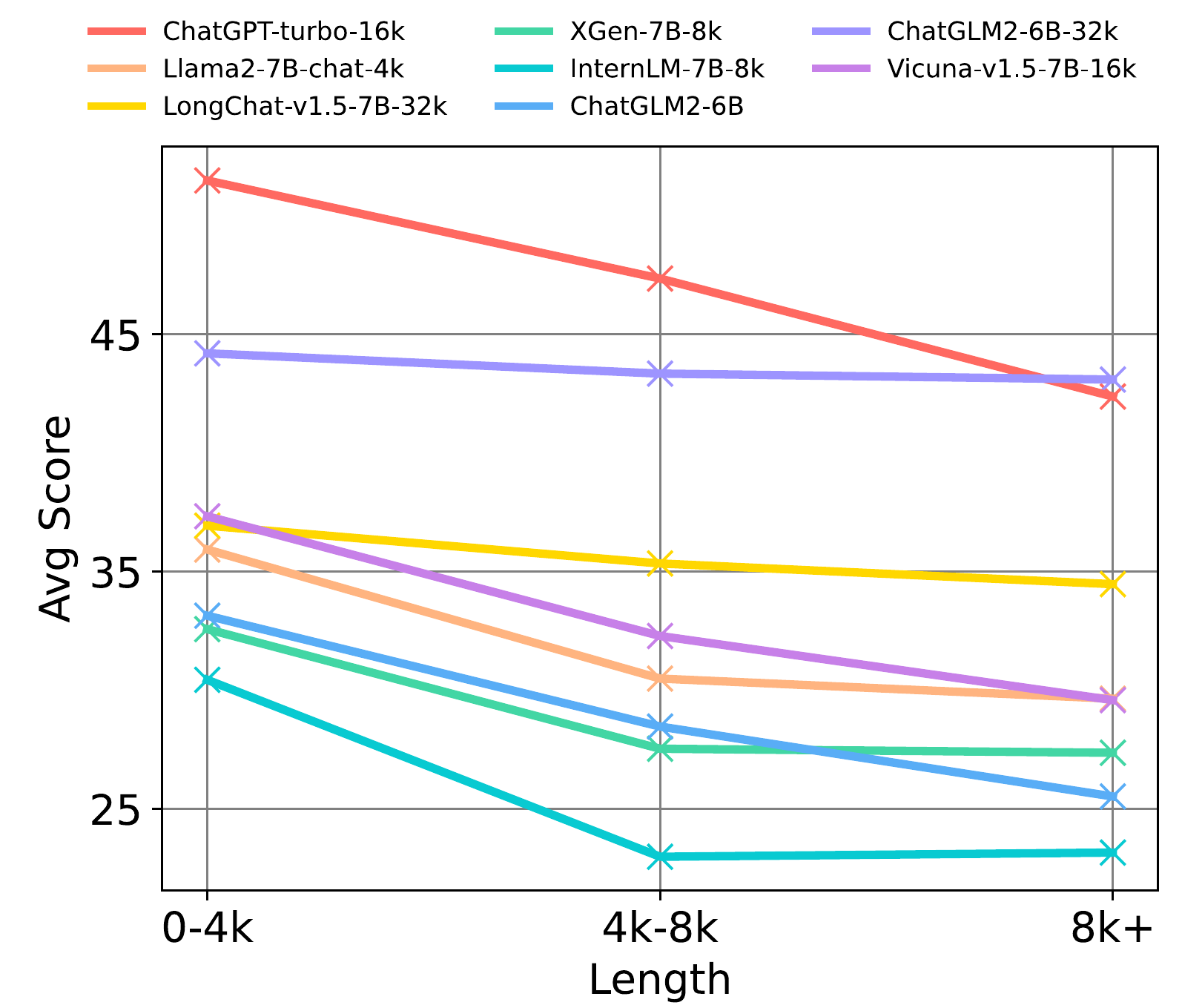}
    \caption{Average score (\%) under different context length on LongBench-E.}
    \label{fig:length}
\end{figure}

\begin{table}[t]
\centering
\resizebox{0.49\textwidth}{!}{
\begin{tabular}{lccccccccc}
\toprule
\multirow{2}{*}{\textbf{Retriever}} & \multicolumn{4}{c}{\textbf{Single-Doc QA}} & \multicolumn{4}{c}{\textbf{Multi-Doc QA}} & \multirow{2}{*}{\textbf{Avg}} \\
\cmidrule(lr){2-5} \cmidrule(lr){6-9}
& \textbf{1-1} & \textbf{1-2} & \textbf{1-3} & \textbf{1-4} & \textbf{2-1} & \textbf{2-2} & \textbf{2-3} & \textbf{2-4} \\
\midrule
\emph{GPT-3.5-Turbo-16k} \\
\rowcolor{mygray}w/o retrieval & {\bf23.6} & {\bf43.3} & 52.3 & {\bf61.2} & {\bf51.6} & 37.7 & 26.9 & 28.7 & {\bf40.7} \\
E-200$\times$7 & 21.8 & 38.1 & 52.8 & 53.6 & 46.6 & {\bf44.9} & {\bf30.4} & {\bf30.7} & 39.9 \\
\rowcolor{mygray}E-500$\times$3 & 21.8 & 39.6 & 50.3 & 55.9 & 49.3 & 38.6 & 23.3 & 30.4 & 38.6 \\
C-200$\times$7 & 18.3 & 35.6 & {\bf54.3} & 52.4 & 47.0 & 39.5 & 25.2 & 30.5 & 37.8 \\
\rowcolor{mygray}C-500$\times$3 & 20.3 & 35.7 & 48.7 & 51.2 & 47.7 & 39.1 & 21.9 & 30.7 & 36.9 \\
B-200$\times$7 & 14.1 & 28.6 & 30.1 & 55.0 & 38.3 & 29.0 & 18.1 & 29.6 & 30.3 \\
\rowcolor{mygray}B-500$\times$3 & 14.5 & 30.4 & 31.3 & 55.1 & 37.2 & 35.1 & 11.7 & 29.9 & 30.6 \\
\midrule
\emph{Llama2-7B-chat-4k} \\
\rowcolor{mygray}w/o retrieval & 18.7 & 19.2 & 36.8 & 11.9 & 25.4 & 32.8 & 9.4 & 5.2 & 19.9 \\
E-200$\times$7 & {\bf20.0} & {\bf25.7} & 40.3 & {\bf13.9} & 34.7 & 34.4 & {\bf17.3} & {\bf5.5} & {\bf24.0} \\
\rowcolor{mygray}E-500$\times$3 & 17.7 & 25.2 & 38.9 & 12.0 & {\bf34.9} & 32.8 & 15.5 & 5.0 & 22.7 \\
C-200$\times$7 & 18.3 & 23.8 & {\bf41.8} & 10.8 & 33.6 & 34.5 & 17.2 & 5.0 & 23.1 \\
\rowcolor{mygray}C-500$\times$3 & 17.1 & 22.5 & 39.5 & 9.9 & 34.6 & {\bf35.0} & 14.1 & 4.7 & 22.2 \\
B-200$\times$7 & 12.3 & 19.6 & 25.9 & 13.1 & 29.2 & 25.9 & 9.1 & 5.1 & 17.5 \\
\rowcolor{mygray}B-500$\times$3 & 14.7 & 20.4 & 26.2 & 13.5 & 23.1 & 29.7 & 7.9 & 5.0 & 17.6 \\
\midrule
\emph{ChatGLM2-6B-32k} \\
\rowcolor{mygray}w/o retrieval & {\bf21.1} & 31.5 & {\bf46.2} & 51.7 & {\bf45.1} & {\bf34.0} & 21.9 & 37.6 & {\bf36.1} \\
E-200$\times$7 & 19.4 & {\bf33.3} & 40.9 & 48.3 & 41.2 & 32.9 & {\bf22.8} & 36.7 & 34.4 \\
\rowcolor{mygray}E-500$\times$3 & 14.6 & 31.2 & 40.5 & 46.3 & 39.4 & 31.5 & 20.2 & {\bf38.1} & 32.7 \\
C-200$\times$7 & 15.1 & 32.9 & 43.1 & 45.8 & 38.3 & 32.3 & 16.9 & 35.5 & 32.5 \\
\rowcolor{mygray}C-500$\times$3 & 12.9 & 29.6 & 41.1 & 49.2 & 38.1 & 33.2 & 17.5 & 37.8 & 32.4 \\
B-200$\times$7 & 12.5 & 20.1 & 23.8 & 50.2 & 28.7 & 24.3 & 10.9 & 35.0 & 25.7 \\
\rowcolor{mygray}B-500$\times$3 & 11.2 & 20.5 & 25.4 & {\bf51.9} & 27.7 & 27.6 & 12.2 & 35.6 & 26.5 \\
\bottomrule
\end{tabular}
}
\caption{Retrieval-based context compression results (\%) on LongBench. E, C, B denote different retrieval methods, namely text-embedding-ada-002, Contriever, and BM25. $M\times N$ indicates the retrieval of the top-$N$ segments when split into chunks by $M$ words. For every model and every dataset, the best performance over all retrieval methods is in \textbf{bold}.}
\label{tb:exp-rtv}
\end{table}

\xhdr{Results on LongBench-E}
While LongBench facilitates the measurement of an overall multi-task ability on tasks that require long context understanding, LongBench-E focuses more on measuring how the model's performance changes under different context lengths within the same task.
As introduced in Sec~\ref{sec:ext}, LongBench-E contains a subset of datasets included in LongBench, featuring more evenly distributed context lengths.
Figure~\ref{fig:length} reports the macro-average scores (\%) on data in length ranges of 0-4k, 4k-8k, and 8k+ (See the results on all datasets in Table~\ref{tb:exp-e}).
One can derive a model's long context ability from the slope of the curve --- 
a significant drop in performance on data of greater length, as indicated by a steeper curve, points to the model's limitations in effectively handling long text modeling.
From the results on LongBench-E, we observe that ChatGLM2-6B-32k and LongChat-v1.5-7B-32k are more robust to longer context length, with a relative drop of 4\% and 7\% from 0-4k to 8k+ respectively.
Moreover, despite GPT-3.5-Turbo-16k demonstrating impressive overall performance across all tasks, we find that it still struggles on longer contexts (-17\% from 0-4k to 8k+), leaving room for future development on long context modeling.

\begin{table}[t]
\centering
\resizebox{0.49\textwidth}{!}{
\begin{tabular}{lcccccc}
\toprule
\textbf{Model} & \textbf{3-1} & \textbf{3-2} & \textbf{3-3} & \textbf{3-4} & \textbf{Avg} \\
\midrule
GPT-3.5-Turbo-16k & 29.5&	23.4&	26.7&	16.0&23.9 \\
\rowcolor{mygray}GPT-3.5-Turbo-16k+Summ & 17.9&	16.6&	17.9&	19.7&18.0\\
Llama2-7B-chat-4k & 27.3&	20.8&	25.8&	0.2&18.5\\
\rowcolor{mygray}Llama2-7B-chat-4k+Summ & 12.8&	16.6&	4.6&	0.6&8.6\\
ChatGLM2-6B-32k & 32.4&	24.0&	26.5&	16.2&24.8\\
\rowcolor{mygray}ChatGLM2-6B-32k+Summ & 17.6&	15.9&	14.9&	17.2&16.4\\
\bottomrule
\end{tabular}
}
\caption{Summarization-based context compression results (\%) on LongBench.}
\label{tb:summ}
\end{table}

\subsection{Impact of Context Compression Techniques}
We further explore the impact of context compression techniques on LongBench, including retrieval-based context compression and summarization-based context compression.
Retrieval is widely used in augmenting language models with external memory~\citep{khandelwal2020generalization,borgeaud2022improving,izacard2022few}.
This application can be extended to consider longer contexts, such as documents or books, as forms of external memory, from which relevant information can be retrieved using a specific query.
Given a long context, we first split it into chunks with a default size of $M$ words (or characters on Chinese datasets), then use a specific retriever to compute the embedding of the text chunks and the query, and concatenate only the top-$N$ chunks according to the cosine similarity of their embeddings to the query embedding.
The top-$N$ chunks as the compressed context, together with the query, are then fed into the model to produce an answer.
A similar pipeline is also implemented in \href{https://python.langchain.com/docs/use_cases/question_answering/}{LangChain}.
We experiment with three retrievers --- OpenAI Embedding (text-embedding-ada-002~\citep{text-embedding-ada-002}), Contriever~\citep{izacard2022unsupervised}, and BM25 --- alongside two chunk sizes of 200 and 500.
In order to conduct a fairer comparison under the same context length, we take the top-7 and top-3 chunks respectively when the chunk sizes are 200 and 500.
Table~\ref{tb:exp-rtv} reports the results on QA tasks in LongBench.
We summarize our findings: (1) text-embedding-ada-002 performs the best among the three retrievers, while the open-sourced Contriever results are closer to text-embedding-ada-002 and superior to BM25. (2) In general, splitting the long context into shorter chunks and retrieving more chunks results in better performance. (3) Under the best retrieval method, the improvements for the three models are -2\%, 21\%, and -5\%, respectively. Moreover, even after retrieval, the performance of Llama2-7B-chat-4k still lags behind the other two models. The results suggest that the retrieval technique can only serve as a performance compensation for models that cannot well model long context, and is not a shortcut to solving long context understanding tasks.

We also study the effect of using model-generated summary as a context compression technique.
Specifically, we first utilize the model to generate a brief summary for each text chunk, and concatenate the summaries together as the compressed context.
We experiment on the summarization tasks in LongBench, and the results are as shown in Table~\ref{tb:summ}.
We find that this compression method improves the performance of the models only on the VCSUM task (3-4), since the data in VCSUM are longer than in the other three datasets.

\begin{table*}[h]
\centering
\resizebox{\textwidth}{!}{
\begin{tabular}{lcccccccc}
\toprule
\textbf{Model} & \textbf{NarrativeQA} & \textbf{Qasper} & \textbf{MultiFieldQA-en} & \textbf{MultiFieldQA-zh} & \textbf{HotpotQA} & \textbf{2WikiMQA} & \textbf{MuSiQue} & \textbf{DuReader} \\
\midrule
GPT-3.5-Turbo-16k (w/o context) & 4.7 & 12.4 & 15.7 & 10.9 & 31.7 & 28.9 & 15.0 & 17.1 \\
GPT-3.5-Turbo-16k & 23.6 (+18.9) & 43.3 (+30.9) & 52.3 (+36.6) & 61.2 (+50.3) & 51.6 (+19.9) & 37.7 (+8.8) & 26.9 (+11.9) & 28.7 (+11.6) \\
\midrule
Llama2-7B-chat-4k (w/o context) & 7.9 & 12.5 & 16.4 & 5.1 & 25.5 & 28.2 & 9.8 & 2.6 \\
Llama2-7B-chat-4k & 18.7 (+10.8) & 19.2 (+6.7) & 36.8 (+20.4) & 11.9 (+6.8) & 25.4 (-0.1) & 32.8 (+4.6) & 9.4 (-0.4) & 5.2 (+2.6) \\
\midrule
ChatGLM2-6B-32k (w/o context) & 8.9 & 14.2 & 12.5 & 20.4 & 17.0 & 19.9 & 7.7 & 16.9 \\
ChatGLM2-6B-32k & 21.1 (+12.2) & 31.5 (+17.3) & 46.2 (+33.7) & 51.6 (+31.2) & 45.1 (+28.1) & 34.0 (+14.1) & 21.9 (+14.2) & 37.6 (+20.7) \\
\bottomrule
\end{tabular}
}
\caption{Assessing context understanding vs. memorization.}
\label{tb:memory}
\end{table*}

\subsection{Context Understanding or Memorization?}
Since the model may have potentially encountered the long context during pre-training, it may rely on memorization rather than context understanding to answer the question.
We conduct an experiment to assess the degree to which these tasks rely on memorization rather than long context understanding capability. Specifically, we withhold the long context from the model, posing only questions to it, and evaluate its performance. Table~\ref{tb:memory} shows the results for GPT-3.5-Turbo-16k, Llama2-7B-chat-4k and ChatGLM2-6B-32k on the Single-Doc QA and Multi-Doc QA datasets in LongBench.

We observe that the memorization performance (w/o context) on HotpotQA, 2WikiMultihopQA, and MusiQue is relatively high. This is likely because these datasets are derived from Wikipedia, a common source for training prevalent LLMs. The $\Delta$ score (original score minus the score when context is absent) addresses the memorization phenomenon~\cite{yu2023kola}, and could also serve as an important indicator for a model's long context understanding ability.

\section{Conclusion}
\label{sec:conclusion}

In this paper, we introduce LongBench, a multi-task bilingual benchmark tailored for gauging long context understanding abilities of LLMs.
LongBench covers six key categories and a total of 21 tasks, with data lengths extending from thousands of tokens up to tens of thousands of tokens.
We also develop LongBench-E which features a more evenly data length distribution.
We conduct extensive experiments on LongBench and LongBench-E, yielding insightful conclusions about the capabilities of current LLMs on long context understanding.
Moreover, our analysis suggests that LongBench and LongBench-E serve as ideal testbeds for future research in long context modeling.

\section{Limitations}
Despite LongBench offers a more comprehensive testbed for long context understanding, it still has its shortcomings, as we summarized below.
(1) Potentially unreliable automatic metrics: As previous studies suggest~\citep{bai2023benchmarking}, the automatic evaluation metrics (ROUGE-L, F1) may not accurately reflect the quality of the response.
Particularly, the results on these metrics may be underestimated for models that are used to generating longer responses.
Although using LLM as examiner may reduce this problem~\citep{bai2023benchmarking,an2023eval}, the runtime overhead for evaluation may be high, and LLM also has bias when used as an evaluation metric~\citep{zheng2023judging}.
(2) Coupling with instruction-following capabilities: Our primary objective is to assess the models' proficiency in long-context modeling regardless of their instruction-following capabilities.
However, as the tasks in LongBench are closer to real-world applications, mastering them inevitably demands a certain level of instruction-following capability.
Consequently, the performance on LongBench is coupled with the models' instruction-following capabilities.

\section*{Acknowledgement}
This work is supported by a grant from the Institute for Guo Qiang, Tsinghua University (2019GQB0003), Zhipu AI and Tsinghua University Initiative Scientific Research Program.
We would also like to thank our anonymous reviewers for their suggestions.

\bibliography{anthology}

\begin{thebibliography}{68}
\expandafter\ifx\csname natexlab\endcsname\relax\def\natexlab#1{#1}\fi

\bibitem[{Ainslie et~al.(2023)Ainslie, Lei, de~Jong, Onta{\~n}{\'o}n, Brahma, Zemlyanskiy, Uthus, Guo, Lee-Thorp, Tay et~al.}]{ainslie2023colt5}
Joshua Ainslie, Tao Lei, Michiel de~Jong, Santiago Onta{\~n}{\'o}n, Siddhartha Brahma, Yury Zemlyanskiy, David Uthus, Mandy Guo, James Lee-Thorp, Yi~Tay, et~al. 2023.
\newblock Colt5: Faster long-range transformers with conditional computation.
\newblock \emph{arXiv preprint arXiv:2303.09752}.

\bibitem[{An et~al.(2023)An, Gong, Zhong, Li, Zhang, Kong, and Qiu}]{an2023eval}
Chenxin An, Shansan Gong, Ming Zhong, Mukai Li, Jun Zhang, Lingpeng Kong, and Xipeng Qiu. 2023.
\newblock L-eval: Instituting standardized evaluation for long context language models.
\newblock \emph{arXiv preprint arXiv:2307.11088}.

\bibitem[{Bai et~al.(2023)Bai, Ying, Cao, Lv, He, Wang, Yu, Zeng, Xiao, Lyu et~al.}]{bai2023benchmarking}
Yushi Bai, Jiahao Ying, Yixin Cao, Xin Lv, Yuze He, Xiaozhi Wang, Jifan Yu, Kaisheng Zeng, Yijia Xiao, Haozhe Lyu, et~al. 2023.
\newblock Benchmarking foundation models with language-model-as-an-examiner.
\newblock \emph{arXiv preprint arXiv:2306.04181}.

\bibitem[{Beltagy et~al.(2020)Beltagy, Peters, and Cohan}]{beltagy2020longformer}
Iz~Beltagy, Matthew~E Peters, and Arman Cohan. 2020.
\newblock Longformer: The long-document transformer.
\newblock \emph{arXiv preprint arXiv:2004.05150}.

\bibitem[{Borgeaud et~al.(2022)Borgeaud, Mensch, Hoffmann, Cai, Rutherford, Millican, Van Den~Driessche, Lespiau, Damoc, Clark et~al.}]{borgeaud2022improving}
Sebastian Borgeaud, Arthur Mensch, Jordan Hoffmann, Trevor Cai, Eliza Rutherford, Katie Millican, George~Bm Van Den~Driessche, Jean-Baptiste Lespiau, Bogdan Damoc, Aidan Clark, et~al. 2022.
\newblock Improving language models by retrieving from trillions of tokens.
\newblock In \emph{International conference on machine learning}, pages 2206--2240. PMLR.

\bibitem[{Bulatov et~al.(2023)Bulatov, Kuratov, and Burtsev}]{bulatov2023scaling}
Aydar Bulatov, Yuri Kuratov, and Mikhail~S Burtsev. 2023.
\newblock Scaling transformer to 1m tokens and beyond with rmt.
\newblock \emph{arXiv preprint arXiv:2304.11062}.

\bibitem[{Bulatov et~al.(2022)Bulatov, Kuratov, and Burtsev}]{bulatov2022recurrent}
Aydar Bulatov, Yury Kuratov, and Mikhail Burtsev. 2022.
\newblock Recurrent memory transformer.
\newblock \emph{Advances in Neural Information Processing Systems}, 35:11079--11091.

\bibitem[{Chen et~al.(2021)Chen, Tworek, Jun, Yuan, Pinto, Kaplan, Edwards, Burda, Joseph, Brockman et~al.}]{chen2021evaluating}
Mark Chen, Jerry Tworek, Heewoo Jun, Qiming Yuan, Henrique Ponde de~Oliveira Pinto, Jared Kaplan, Harri Edwards, Yuri Burda, Nicholas Joseph, Greg Brockman, et~al. 2021.
\newblock Evaluating large language models trained on code.
\newblock \emph{arXiv preprint arXiv:2107.03374}.

\bibitem[{Chen et~al.(2023)Chen, Wong, Chen, and Tian}]{chen2023extending}
Shouyuan Chen, Sherman Wong, Liangjian Chen, and Yuandong Tian. 2023.
\newblock Extending context window of large language models via positional interpolation.
\newblock \emph{arXiv preprint arXiv:2306.15595}.

\bibitem[{Child et~al.(2019)Child, Gray, Radford, and Sutskever}]{child2019generating}
Rewon Child, Scott Gray, Alec Radford, and Ilya Sutskever. 2019.
\newblock Generating long sequences with sparse transformers.
\newblock \emph{arXiv preprint arXiv:1904.10509}.

\bibitem[{Dai et~al.(2019)Dai, Yang, Yang, Carbonell, Le, and Salakhutdinov}]{dai2019transformer}
Zihang Dai, Zhilin Yang, Yiming Yang, Jaime~G Carbonell, Quoc Le, and Ruslan Salakhutdinov. 2019.
\newblock Transformer-xl: Attentive language models beyond a fixed-length context.
\newblock In \emph{Proceedings of the 57th Annual Meeting of the Association for Computational Linguistics}, pages 2978--2988.

\bibitem[{Dasigi et~al.(2021)Dasigi, Lo, Beltagy, Cohan, Smith, and Gardner}]{dasigi2021dataset}
Pradeep Dasigi, Kyle Lo, Iz~Beltagy, Arman Cohan, Noah~A Smith, and Matt Gardner. 2021.
\newblock A dataset of information-seeking questions and answers anchored in research papers.
\newblock In \emph{Proceedings of the 2021 Conference of the North American Chapter of the Association for Computational Linguistics: Human Language Technologies}, pages 4599--4610.

\bibitem[{Ding et~al.(2023)Ding, Ma, Dong, Zhang, Huang, Wang, and Wei}]{ding2023longnet}
Jiayu Ding, Shuming Ma, Li~Dong, Xingxing Zhang, Shaohan Huang, Wenhui Wang, and Furu Wei. 2023.
\newblock Longnet: Scaling transformers to 1,000,000,000 tokens.
\newblock \emph{arXiv preprint arXiv:2307.02486}.

\bibitem[{Du et~al.(2022)Du, Qian, Liu, Ding, Qiu, Yang, and Tang}]{du2022glm}
Zhengxiao Du, Yujie Qian, Xiao Liu, Ming Ding, Jiezhong Qiu, Zhilin Yang, and Jie Tang. 2022.
\newblock Glm: General language model pretraining with autoregressive blank infilling.
\newblock In \emph{Proceedings of the 60th Annual Meeting of the Association for Computational Linguistics (Volume 1: Long Papers)}, pages 320--335.

\bibitem[{Fabbri et~al.(2019)Fabbri, Li, She, Li, and Radev}]{fabbri2019multi}
Alexander~Richard Fabbri, Irene Li, Tianwei She, Suyi Li, and Dragomir Radev. 2019.
\newblock Multi-news: A large-scale multi-document summarization dataset and abstractive hierarchical model.
\newblock In \emph{Proceedings of the 57th Annual Meeting of the Association for Computational Linguistics}, pages 1074--1084.

\bibitem[{Fedus et~al.(2022)Fedus, Zoph, and Shazeer}]{fedus2022switch}
William Fedus, Barret Zoph, and Noam Shazeer. 2022.
\newblock Switch transformers: Scaling to trillion parameter models with simple and efficient sparsity.
\newblock \emph{The Journal of Machine Learning Research}, 23(1):5232--5270.

\bibitem[{Gliwa et~al.(2019)Gliwa, Mochol, Biesek, and Wawer}]{gliwa2019samsum}
Bogdan Gliwa, Iwona Mochol, Maciej Biesek, and Aleksander Wawer. 2019.
\newblock Samsum corpus: A human-annotated dialogue dataset for abstractive summarization.
\newblock \emph{EMNLP-IJCNLP 2019}, page~70.

\bibitem[{Guo et~al.(2023)Guo, Xu, Duan, Yin, and McAuley}]{guo2023longcoder}
Daya Guo, Canwen Xu, Nan Duan, Jian Yin, and Julian McAuley. 2023.
\newblock Longcoder: A long-range pre-trained language model for code completion.
\newblock \emph{arXiv preprint arXiv:2306.14893}.

\bibitem[{He et~al.(2018)He, Liu, Liu, Lyu, Zhao, Xiao, Liu, Wang, Wu, She et~al.}]{he2018dureader}
Wei He, Kai Liu, Jing Liu, Yajuan Lyu, Shiqi Zhao, Xinyan Xiao, Yuan Liu, Yizhong Wang, Hua Wu, Qiaoqiao She, et~al. 2018.
\newblock Dureader: a chinese machine reading comprehension dataset from real-world applications.
\newblock \emph{ACL 2018}, page~37.

\bibitem[{Hendrycks et~al.(2021)Hendrycks, Burns, Basart, Zou, Mazeika, Song, and Steinhardt}]{hendrycks2021measuring}
Dan Hendrycks, Collin Burns, Steven Basart, Andy Zou, Mantas Mazeika, Dawn Song, and Jacob Steinhardt. 2021.
\newblock Measuring massive multitask language understanding.
\newblock In \emph{International Conference on Learning Representations}.

\bibitem[{Ho et~al.(2020)Ho, Nguyen, Sugawara, and Aizawa}]{ho2020constructing}
Xanh Ho, Anh-Khoa~Duong Nguyen, Saku Sugawara, and Akiko Aizawa. 2020.
\newblock Constructing a multi-hop qa dataset for comprehensive evaluation of reasoning steps.
\newblock In \emph{Proceedings of the 28th International Conference on Computational Linguistics}, pages 6609--6625.

\bibitem[{Huang et~al.(2021)Huang, Cao, Parulian, Ji, and Wang}]{huang2021efficient}
Luyang Huang, Shuyang Cao, Nikolaus Parulian, Heng Ji, and Lu~Wang. 2021.
\newblock Efficient attentions for long document summarization.
\newblock In \emph{Proceedings of the 2021 Conference of the North American Chapter of the Association for Computational Linguistics: Human Language Technologies}, pages 1419--1436.

\bibitem[{Izacard et~al.(2022{\natexlab{a}})Izacard, Caron, Hosseini, Riedel, Bojanowski, Joulin, and Grave}]{izacard2022unsupervised}
Gautier Izacard, Mathilde Caron, Lucas Hosseini, Sebastian Riedel, Piotr Bojanowski, Armand Joulin, and Edouard Grave. 2022{\natexlab{a}}.
\newblock Unsupervised dense information retrieval with contrastive learning.
\newblock \emph{Transactions on Machine Learning Research}.

\bibitem[{Izacard et~al.(2022{\natexlab{b}})Izacard, Lewis, Lomeli, Hosseini, Petroni, Schick, Dwivedi-Yu, Joulin, Riedel, and Grave}]{izacard2022few}
Gautier Izacard, Patrick Lewis, Maria Lomeli, Lucas Hosseini, Fabio Petroni, Timo Schick, Jane Dwivedi-Yu, Armand Joulin, Sebastian Riedel, and Edouard Grave. 2022{\natexlab{b}}.
\newblock Few-shot learning with retrieval augmented language models.
\newblock \emph{arXiv preprint arXiv:2208.03299}.

\bibitem[{Joshi et~al.(2017)Joshi, Choi, Weld, and Zettlemoyer}]{joshi2017triviaqa}
Mandar Joshi, Eunsol Choi, Daniel~S Weld, and Luke Zettlemoyer. 2017.
\newblock Triviaqa: A large scale distantly supervised challenge dataset for reading comprehension.
\newblock In \emph{Proceedings of the 55th Annual Meeting of the Association for Computational Linguistics (Volume 1: Long Papers)}, pages 1601--1611.

\bibitem[{Khandelwal et~al.(2020)Khandelwal, Levy, Jurafsky, Zettlemoyer, and Lewis}]{khandelwal2020generalization}
Urvashi Khandelwal, Omer Levy, Dan Jurafsky, Luke Zettlemoyer, and Mike Lewis. 2020.
\newblock Generalization through memorization: Nearest neighbor language models.
\newblock In \emph{International Conference on Learning Representations}.

\bibitem[{Kitaev et~al.(2020)Kitaev, Kaiser, and Levskaya}]{kitaev2020reformer}
Nikita Kitaev, Lukasz Kaiser, and Anselm Levskaya. 2020.
\newblock Reformer: The efficient transformer.
\newblock In \emph{International Conference on Learning Representations}.

\bibitem[{Ko{\v{c}}isk{\`y} et~al.(2018)Ko{\v{c}}isk{\`y}, Schwarz, Blunsom, Dyer, Hermann, Melis, and Grefenstette}]{kovcisky2018narrativeqa}
Tom{\'a}{\v{s}} Ko{\v{c}}isk{\`y}, Jonathan Schwarz, Phil Blunsom, Chris Dyer, Karl~Moritz Hermann, G{\'a}bor Melis, and Edward Grefenstette. 2018.
\newblock The narrativeqa reading comprehension challenge.
\newblock \emph{Transactions of the Association for Computational Linguistics}, 6:317--328.

\bibitem[{Li et~al.(2023)Li, Shao, Xie, Sheng, Zheng, Gonzalez, Stoica, Ma, and Zhang}]{longchat2023}
Dacheng Li, Rulin Shao, Anze Xie, Ying Sheng, Lianmin Zheng, Joseph~E. Gonzalez, Ion Stoica, Xuezhe Ma, and Hao Zhang. 2023.
\newblock \href {https://lmsys.org/blog/2023-06-29-longchat} {How long can open-source llms truly promise on context length?}

\bibitem[{Li and Roth(2002)}]{li2002learning}
Xin Li and Dan Roth. 2002.
\newblock Learning question classifiers.
\newblock In \emph{COLING 2002: The 19th International Conference on Computational Linguistics}.

\bibitem[{Liang et~al.(2023)Liang, Wang, Huang, Wu, Wu, Lu, Ma, and Li}]{liang2023unleashing}
Xinnian Liang, Bing Wang, Hui Huang, Shuangzhi Wu, Peihao Wu, Lu~Lu, Zejun Ma, and Zhoujun Li. 2023.
\newblock Unleashing infinite-length input capacity for large-scale language models with self-controlled memory system.
\newblock \emph{arXiv preprint arXiv:2304.13343}.

\bibitem[{Lin(2004)}]{lin2004rouge}
Chin-Yew Lin. 2004.
\newblock Rouge: A package for automatic evaluation of summaries.
\newblock In \emph{Text summarization branches out}, pages 74--81.

\bibitem[{Liu et~al.(2023{\natexlab{a}})Liu, Lin, Hewitt, Paranjape, Bevilacqua, Petroni, and Liang}]{liu2023lost}
Nelson~F Liu, Kevin Lin, John Hewitt, Ashwin Paranjape, Michele Bevilacqua, Fabio Petroni, and Percy Liang. 2023{\natexlab{a}}.
\newblock Lost in the middle: How language models use long contexts.
\newblock \emph{arXiv preprint arXiv:2307.03172}.

\bibitem[{Liu et~al.(2023{\natexlab{b}})Liu, Xu, and McAuley}]{liu2023repobench}
Tianyang Liu, Canwen Xu, and Julian McAuley. 2023{\natexlab{b}}.
\newblock Repobench: Benchmarking repository-level code auto-completion systems.
\newblock \emph{arXiv preprint arXiv:2306.03091}.

\bibitem[{Liu et~al.(2023{\natexlab{c}})Liu, Yu, Zhang, Xu, Lei, Lai, Gu, Ding, Men, Yang et~al.}]{liu2023agentbench}
Xiao Liu, Hao Yu, Hanchen Zhang, Yifan Xu, Xuanyu Lei, Hanyu Lai, Yu~Gu, Hangliang Ding, Kaiwen Men, Kejuan Yang, et~al. 2023{\natexlab{c}}.
\newblock Agentbench: Evaluating llms as agents.
\newblock \emph{arXiv preprint arXiv:2308.03688}.

\bibitem[{Martins et~al.(2022)Martins, Marinho, and Martins}]{martins2022former}
Pedro~Henrique Martins, Zita Marinho, and Andr{\'e}~FT Martins. 2022.
\newblock $\infty$-former: Infinite memory transformer.
\newblock In \emph{Proceedings of the 60th Annual Meeting of the Association for Computational Linguistics (Volume 1: Long Papers)}, pages 5468--5485.

\bibitem[{Mohtashami and Jaggi(2023)}]{mohtashami2023landmark}
Amirkeivan Mohtashami and Martin Jaggi. 2023.
\newblock Landmark attention: Random-access infinite context length for transformers.
\newblock \emph{arXiv preprint arXiv:2305.16300}.

\bibitem[{Nijkamp et~al.(2023)Nijkamp, Xie, Hayashi, Pang, Xia, Xing, Vig, Yavuz, Laban et~al.}]{XGen}
Erik Nijkamp, Tian Xie, Hiroaki Hayashi, Bo~Pang, Congying Xia, Chen Xing, Jesse Vig, Semih Yavuz, Philippe Laban, et~al. 2023.
\newblock \href {https://blog.salesforceairesearch.com/xgen} {Long sequence modeling with xgen: A 7b llm trained on 8k input sequence length}.
\newblock Salesforce AI Research Blog.

\bibitem[{NLPCC(2014)}]{lsht}
NLPCC. 2014.
\newblock \href {http://tcci.ccf.org.cn/conference/2014/dldoc/evatask6.pdf} {Task definition for large scale text categorization at nlpcc 2014}.

\bibitem[{OpenAI(2022{\natexlab{a}})}]{chatgpt}
OpenAI. 2022{\natexlab{a}}.
\newblock \href {https://openai.com/blog/chatgpt} {Introducing chatgpt}.

\bibitem[{OpenAI(2022{\natexlab{b}})}]{text-embedding-ada-002}
OpenAI. 2022{\natexlab{b}}.
\newblock \href {https://openai.com/blog/new-and-improved-embedding-model} {Openai: New and improved embedding model}.

\bibitem[{Orvieto et~al.(2023)Orvieto, Smith, Gu, Fernando, Gulcehre, Pascanu, and De}]{orvieto2023resurrecting}
Antonio Orvieto, Samuel~L Smith, Albert Gu, Anushan Fernando, Caglar Gulcehre, Razvan Pascanu, and Soham De. 2023.
\newblock Resurrecting recurrent neural networks for long sequences.
\newblock \emph{arXiv preprint arXiv:2303.06349}.

\bibitem[{Press et~al.(2022)Press, Smith, and Lewis}]{press2022train}
Ofir Press, Noah Smith, and Mike Lewis. 2022.
\newblock Train short, test long: Attention with linear biases enables input length extrapolation.
\newblock In \emph{International Conference on Learning Representations}.

\bibitem[{Rae et~al.(2020)Rae, Potapenko, Jayakumar, Hillier, and Lillicrap}]{rae2020compressive}
Jack~W Rae, Anna Potapenko, Siddhant~M Jayakumar, Chloe Hillier, and Timothy~P Lillicrap. 2020.
\newblock Compressive transformers for long-range sequence modelling.
\newblock In \emph{International Conference on Learning Representations}.

\bibitem[{Raffel et~al.(2020)Raffel, Shazeer, Roberts, Lee, Narang, Matena, Zhou, Li, and Liu}]{t5}
Colin Raffel, Noam Shazeer, Adam Roberts, Katherine Lee, Sharan Narang, Michael Matena, Yanqi Zhou, Wei Li, and Peter~J Liu. 2020.
\newblock Exploring the limits of transfer learning with a unified text-to-text transformer.
\newblock \emph{The Journal of Machine Learning Research}, 21(1):5485--5551.

\bibitem[{Roy et~al.(2021)Roy, Saffar, Vaswani, and Grangier}]{roy2021efficient}
Aurko Roy, Mohammad Saffar, Ashish Vaswani, and David Grangier. 2021.
\newblock Efficient content-based sparse attention with routing transformers.
\newblock \emph{Transactions of the Association for Computational Linguistics}, 9:53--68.

\bibitem[{Shaham et~al.(2023)Shaham, Ivgi, Efrat, Berant, and Levy}]{shaham2023zeroscrolls}
Uri Shaham, Maor Ivgi, Avia Efrat, Jonathan Berant, and Omer Levy. 2023.
\newblock Zeroscrolls: A zero-shot benchmark for long text understanding.
\newblock \emph{arXiv preprint arXiv:2305.14196}.

\bibitem[{Shaham et~al.(2022)Shaham, Segal, Ivgi, Efrat, Yoran, Haviv, Gupta, Xiong, Geva, Berant et~al.}]{shaham2022scrolls}
Uri Shaham, Elad Segal, Maor Ivgi, Avia Efrat, Ori Yoran, Adi Haviv, Ankit Gupta, Wenhan Xiong, Mor Geva, Jonathan Berant, et~al. 2022.
\newblock Scrolls: Standardized comparison over long language sequences.
\newblock In \emph{Proceedings of the 2022 Conference on Empirical Methods in Natural Language Processing}, pages 12007--12021.

\bibitem[{Srivastava et~al.(2023)Srivastava, Rastogi, Rao, Shoeb, Abid, Fisch, Brown, Santoro, Gupta, Garriga-Alonso et~al.}]{srivastava2023beyond}
Aarohi Srivastava, Abhinav Rastogi, Abhishek Rao, Abu Awal~Md Shoeb, Abubakar Abid, Adam Fisch, Adam~R Brown, Adam Santoro, Aditya Gupta, Adri{\`a} Garriga-Alonso, et~al. 2023.
\newblock Beyond the imitation game: Quantifying and extrapolating the capabilities of language models.
\newblock \emph{Transactions on Machine Learning Research}.

\bibitem[{Sun et~al.(2021)Sun, Krishna, Mattarella-Micke, and Iyyer}]{sun2021long}
Simeng Sun, Kalpesh Krishna, Andrew Mattarella-Micke, and Mohit Iyyer. 2021.
\newblock Do long-range language models actually use long-range context?
\newblock In \emph{Proceedings of the 2021 Conference on Empirical Methods in Natural Language Processing}, pages 807--822.

\bibitem[{Sun et~al.(2022)Sun, Dong, Patra, Ma, Huang, Benhaim, Chaudhary, Song, and Wei}]{sun2022length}
Yutao Sun, Li~Dong, Barun Patra, Shuming Ma, Shaohan Huang, Alon Benhaim, Vishrav Chaudhary, Xia Song, and Furu Wei. 2022.
\newblock A length-extrapolatable transformer.
\newblock \emph{arXiv preprint arXiv:2212.10554}.

\bibitem[{Svyatkovskiy et~al.(2020)Svyatkovskiy, Deng, Fu, and Sundaresan}]{svyatkovskiy2020intellicode}
Alexey Svyatkovskiy, Shao~Kun Deng, Shengyu Fu, and Neel Sundaresan. 2020.
\newblock Intellicode compose: Code generation using transformer.
\newblock In \emph{Proceedings of the 28th ACM Joint Meeting on European Software Engineering Conference and Symposium on the Foundations of Software Engineering}, pages 1433--1443.

\bibitem[{Tay et~al.(2021)Tay, Dehghani, Abnar, Shen, Bahri, Pham, Rao, Yang, Ruder, and Metzler}]{tay2021long}
Yi~Tay, Mostafa Dehghani, Samira Abnar, Yikang Shen, Dara Bahri, Philip Pham, Jinfeng Rao, Liu Yang, Sebastian Ruder, and Donald Metzler. 2021.
\newblock Long range arena: A benchmark for efficient transformers.
\newblock In \emph{International Conference on Learning Representations}.

\bibitem[{Tay et~al.(2022)Tay, Dehghani, Bahri, and Metzler}]{tay2022efficient}
Yi~Tay, Mostafa Dehghani, Dara Bahri, and Donald Metzler. 2022.
\newblock \href {https://doi.org/10.1145/3530811} {Efficient transformers: A survey}.
\newblock \emph{ACM Comput. Surv.}, 55(6).

\bibitem[{Team(2023)}]{2023internlm}
InternLM Team. 2023.
\newblock Internlm: A multilingual language model with progressively enhanced capabilities.
\newblock \url{https://github.com/InternLM/InternLM}.

\bibitem[{Touvron et~al.(2023)Touvron, Martin, Stone, Albert, Almahairi, Babaei, Bashlykov, Batra, Bhargava, Bhosale et~al.}]{touvron2023llama}
Hugo Touvron, Louis Martin, Kevin Stone, Peter Albert, Amjad Almahairi, Yasmine Babaei, Nikolay Bashlykov, Soumya Batra, Prajjwal Bhargava, Shruti Bhosale, et~al. 2023.
\newblock Llama 2: Open foundation and fine-tuned chat models.
\newblock \emph{arXiv preprint arXiv:2307.09288}.

\bibitem[{Trivedi et~al.(2022)Trivedi, Balasubramanian, Khot, and Sabharwal}]{trivedi2022musique}
Harsh Trivedi, Niranjan Balasubramanian, Tushar Khot, and Ashish Sabharwal. 2022.
\newblock ♫ musique: Multihop questions via single-hop question composition.
\newblock \emph{Transactions of the Association for Computational Linguistics}, 10:539--554.

\bibitem[{Wang et~al.(2020)Wang, Li, Khabsa, Fang, and Ma}]{wang2020linformer}
Sinong Wang, Belinda~Z Li, Madian Khabsa, Han Fang, and Hao Ma. 2020.
\newblock Linformer: Self-attention with linear complexity.
\newblock \emph{arXiv preprint arXiv:2006.04768}.

\bibitem[{Wu et~al.(2023)Wu, Zhan, Tan, Hou, Liang, and Song}]{wu-etal-2023-vcsum}
Han Wu, Mingjie Zhan, Haochen Tan, Zhaohui Hou, Ding Liang, and Linqi Song. 2023.
\newblock {VCSUM}: A versatile {C}hinese meeting summarization dataset.
\newblock In \emph{Findings of the Association for Computational Linguistics: ACL 2023}, pages 6065--6079.

\bibitem[{Wu et~al.(2022)Wu, Rabe, Hutchins, and Szegedy}]{wu2022memorizing}
Yuhuai Wu, Markus~Norman Rabe, DeLesley Hutchins, and Christian Szegedy. 2022.
\newblock Memorizing transformers.
\newblock In \emph{International Conference on Learning Representations}.

\bibitem[{Yang et~al.(2018)Yang, Qi, Zhang, Bengio, Cohen, Salakhutdinov, and Manning}]{yang2018hotpotqa}
Zhilin Yang, Peng Qi, Saizheng Zhang, Yoshua Bengio, William Cohen, Ruslan Salakhutdinov, and Christopher~D Manning. 2018.
\newblock Hotpotqa: A dataset for diverse, explainable multi-hop question answering.
\newblock In \emph{Proceedings of the 2018 Conference on Empirical Methods in Natural Language Processing}, pages 2369--2380.

\bibitem[{Yu et~al.(2024)Yu, Wang, Tu, Cao, Zhang-Li, Lv, Peng, Yao, Zhang, Li et~al.}]{yu2023kola}
Jifan Yu, Xiaozhi Wang, Shangqing Tu, Shulin Cao, Daniel Zhang-Li, Xin Lv, Hao Peng, Zijun Yao, Xiaohan Zhang, Hanming Li, et~al. 2024.
\newblock Kola: Carefully benchmarking world knowledge of large language models.
\newblock In \emph{The Twelfth International Conference on Learning Representations}.

\bibitem[{Zaheer et~al.(2020)Zaheer, Guruganesh, Dubey, Ainslie, Alberti, Ontanon, Pham, Ravula, Wang, Yang et~al.}]{zaheer2020big}
Manzil Zaheer, Guru Guruganesh, Kumar~Avinava Dubey, Joshua Ainslie, Chris Alberti, Santiago Ontanon, Philip Pham, Anirudh Ravula, Qifan Wang, Li~Yang, et~al. 2020.
\newblock Big bird: Transformers for longer sequences.
\newblock \emph{Advances in neural information processing systems}, 33:17283--17297.

\bibitem[{Zeng et~al.(2023)Zeng, Liu, Du, Wang, Lai, Ding, Yang, Xu, Zheng, Xia et~al.}]{zeng2022glm}
Aohan Zeng, Xiao Liu, Zhengxiao Du, Zihan Wang, Hanyu Lai, Ming Ding, Zhuoyi Yang, Yifan Xu, Wendi Zheng, Xiao Xia, et~al. 2023.
\newblock Glm-130b: An open bilingual pre-trained model.
\newblock In \emph{The Eleventh International Conference on Learning Representations}.

\bibitem[{Zheng et~al.(2023{\natexlab{a}})Zheng, Chiang, Sheng, Zhuang, Wu, Zhuang, Lin, Li, Li, Xing et~al.}]{zheng2023judging}
Lianmin Zheng, Wei-Lin Chiang, Ying Sheng, Siyuan Zhuang, Zhanghao Wu, Yonghao Zhuang, Zi~Lin, Zhuohan Li, Dacheng Li, Eric Xing, et~al. 2023{\natexlab{a}}.
\newblock Judging llm-as-a-judge with mt-bench and chatbot arena.
\newblock \emph{arXiv preprint arXiv:2306.05685}.

\bibitem[{Zheng et~al.(2023{\natexlab{b}})Zheng, Xia, Zou, Dong, Wang, Xue, Shen, Wang, Wang, Li et~al.}]{zheng2023codegeex}
Qinkai Zheng, Xiao Xia, Xu~Zou, Yuxiao Dong, Shan Wang, Yufei Xue, Lei Shen, Zihan Wang, Andi Wang, Yang Li, et~al. 2023{\natexlab{b}}.
\newblock Codegeex: A pre-trained model for code generation with multilingual benchmarking on humaneval-x.
\newblock In \emph{Proceedings of the 29th ACM SIGKDD Conference on Knowledge Discovery and Data Mining}, pages 5673--5684.

\bibitem[{Zhong et~al.(2021)Zhong, Yin, Yu, Zaidi, Mutuma, Jha, Hassan, Celikyilmaz, Liu, Qiu et~al.}]{zhong2021qmsum}
Ming Zhong, Da~Yin, Tao Yu, Ahmad Zaidi, Mutethia Mutuma, Rahul Jha, Ahmed Hassan, Asli Celikyilmaz, Yang Liu, Xipeng Qiu, et~al. 2021.
\newblock Qmsum: A new benchmark for query-based multi-domain meeting summarization.
\newblock In \emph{Proceedings of the 2021 Conference of the North American Chapter of the Association for Computational Linguistics: Human Language Technologies}, pages 5905--5921.

\bibitem[{Zhou et~al.(2023)Zhou, Jiang, Cui, Wang, Xiao, Hou, Cotterell, and Sachan}]{zhou2023recurrentgpt}
Wangchunshu Zhou, Yuchen~Eleanor Jiang, Peng Cui, Tiannan Wang, Zhenxin Xiao, Yifan Hou, Ryan Cotterell, and Mrinmaya Sachan. 2023.
\newblock Recurrentgpt: Interactive generation of (arbitrarily) long text.
\newblock \emph{arXiv preprint arXiv:2305.13304}.

\end{thebibliography}

\appendix
\newpage
\appendix
\onecolumn
\section{Dataset Details}
Table~\ref{tb:form} lists the instantiation of $(I, C, A)$ for each dataset in LongBench.
Table~\ref{tb:e} reports the number of data on each task that falls in the length range of 0-4k, 4k-8k, and 8k+ in LongBench-E.

\begin{table}[htbp]
\centering  
\resizebox{0.8\textwidth}{!}{
\begin{tabular}{llll}
\toprule
Dataset & Input $I$ & Context $C$ & Answer $A$ \\
\midrule
\emph{Single-Document QA} \\
NarrativeQA & Question & Document & Answer \\
Qasper & Question & Document & Answer \\
MultiFieldQA-en & Question & Document & Answer \\
MultiFieldQA-zh & Question & Document & Answer \\
\midrule
\emph{Multi-Document QA} \\
HotpotQA & Question & Multiple documents & Answer \\
2WikiMultihopQA & Question & Multiple documents & Answer \\
MuSiQue & Question & Multiple documents & Answer \\
DuReader & Question & Multiple documents & Answer \\
\midrule
\emph{Summarization} \\
GovReport & - & Document & Summary \\
QMSum & Query & Document & Summary \\
MultiNews & - & Document & Summary \\
VCSUM & - & Document & Summary \\
\midrule
\emph{Few-shot Learning} \\
TREC & Question & Few-shot examples & Class label \\
TriviaQA & Passage\&Question & Few-shot examples & Answer \\
SAMSum & Dialogue & Few-shot examples & Summary  \\
LSHT & News document & Few-shot examples & Class label \\
\midrule
\emph{Synthetic Task} \\
PassageCount & - & Multiple passages & Count \\
PassageRetrieval-en & Summary & Multiple passages & Title of the passage \\
PassageRetrieval-zh & Summary & Multiple passages & Title of the passage \\
\midrule
\emph{Code Completion} \\
LCC & - & Preceding lines of code & Next line of code \\
RepoBench-P & Preceding lines of code & Cross-file code snippets & Next line of code \\
\bottomrule
\end{tabular}
}
\caption{Instantiation of $(I, C, A)$ for each task in LongBench.}
\label{tb:form}
\end{table}

\begin{table}[htbp]
\centering  
\begin{tabular}{lccc}
\toprule
Dataset & \#data in 0-4k & \#data in 4-8k & \#data in 8k+ \\
\midrule
\emph{Single-Document QA} \\
Qasper & 100 & 100 & 24 \\
MultiFieldQA-en & 67 & 70 & 13 \\
\midrule
\emph{Multi-Document QA} \\
HotpotQA & 100 & 100 & 100 \\
2WikiMultihopQA & 100 & 100 & 100 \\
\midrule
\emph{Summarization} \\
GovReport & 100 & 100 & 100 \\
MultiNews & 100 & 100 & 94 \\
\midrule
\emph{Few-shot Learning} \\
TREC & 100 & 100 & 100 \\
TriviaQA & 100 & 100 & 100 \\
SAMSum & 100 & 100 & 100  \\
\midrule
\emph{Synthetic Task} \\
PassageCount & 100 & 100 & 100 \\
PassageRetrieval-en & 100 & 100 & 100 \\
\midrule
\emph{Code Completion} \\
LCC & 100 & 100 & 100 \\
RepoBench-P & 100 & 100 & 100 \\
\bottomrule
\end{tabular}
\caption{Data length distributions in LongBench-E.}
\label{tb:e}
\end{table}

\xhdr{Details of MultiFieldQA document sources}
The source of the documents in MultiFieldQA include:
\begin{itemize}
    \item \href{https://arxiv.org/}{Arxiv} (for academic papers): open-accessed and can be downloaded freely by anyone.
    \item \href{https://huggingface.co/datasets/allenai/c4}{C4 Dataset}: publicly available dataset with ODC-BY license.
    \item \href{https://data.baai.ac.cn/details/WuDaoCorporaText}{WuDaoCorpora}: open-accessed dataset.
    \item \href{https://wenshu.court.gov.cn/}{Chinese Judgements Online} (for Chinese legal documents): open Chinese judgements download website.
    \item \href{https://www.wikipedia.org/}{Wikipedia} (for encyclopedias): grant free access and licensed under CC BY-SA.
    \item \href{https://www.gov.cn/}{Chinese Government Website} (for Chinese government report): open Chinese government report download website.
\end{itemize}

\xhdr{Annotation guidelines for MultiFieldQA}
Here we provide the annotation guidelines: ``\emph{Please propose one question and a groundtruth answer for each of the following documents. Requirements: 1. Questions should be as clear as possible and have definitive and relatively short answers. 2. The distribution of the evidence paragraphs should be as random as possible throughout the article. 3. Ensure that the questions have varied types, including, but not limited to, information-extracting questions (e.g., the time of an event, the date of birth of a person, etc.), summarizing questions (e.g., which people do the article mainly describes), and multi-hop reasoning questions.}''

The average time taken to annotate each data sample in MultiFieldQA is around 5 minutes. The annotators involved in this process are Ph.D. students with extensive research experience in the field of NLP, positioning them as expert annotators.
Cross-validation among annotators shows a 100\% accuracy rate of the annotated answers.

\section{Evaluation Setups}
\xhdr{Evaluation Prompts}
In this section, we present a collection of customized prompt templates designed for each dataset within LongBench, utilized during our evaluation. Recall that each data instance is accompanied by an input $I$ as well as a context $C$. 
We place the instruction both at the beginning and end of the prompt, ensuring the models fully grasp what to do.

\begin{tcolorbox}[size=title,opacityfill=0.1,breakable]
\noindent
\textbf{NarrativeQA}: You are given a story, which can be either a novel or a movie script, and a question. Answer the question as concisely as you can, using a single phrase if possible. Do not provide any explanation.\\
Story: \{context\}\\
Now, answer the question based on the story as concisely as you can, using a single phrase if possible. Do not provide any explanation.\\
Question: \{input\}\\
Answer:
\end{tcolorbox}

\begin{tcolorbox}[size=title,opacityfill=0.1,breakable]
\noindent
\textbf{Qasper}: You are given a scientific article and a question. Answer the question as concisely as you can, using a single phrase or sentence if possible. If the question cannot be answered based on the information in the article, write ``unanswerable''. If the question is a yes/no question, answer ``yes'', ``no'', or ``unanswerable''. Do not provide any explanation.\\
Article: \{context\}\\
Answer the question based on the above article as concisely as you can, using a single phrase or sentence if possible. If the question cannot be answered based on the information in the article, write ``unanswerable''. If the question is a yes/no question, answer ``yes'', ``no'', or ``unanswerable''. Do not provide any explanation.\\
Question: \{input\}\\
Answer:
\end{tcolorbox}

\begin{tcolorbox}[size=title,opacityfill=0.1,breakable]
\noindent
\textbf{MultiField-en}: Read the following text and answer briefly.\\
\{context\}\\
Now, answer the following question based on the above text, only give me the answer and do not output any other words.\\
Question: \{input\}\\
Answer:
\end{tcolorbox}

\begin{tcolorbox}[size=title,opacityfill=0.1,breakable]
\noindent
\textbf{MultiField-zh}: 
\begin{CJK}{UTF8}{gbsn}
阅读以下文字并用中文简短回答：\\
\{context\}\\
现在请基于上面的文章回答下面的问题，只告诉我答案，不要输出任何其他字词。\\
问题：\{input\}\\
回答：
\end{CJK}
\end{tcolorbox}

\begin{tcolorbox}[size=title,opacityfill=0.1,breakable]
\noindent
\textbf{HotpotQA}: Answer the question based on the given passages. Only give me the answer and do not output any other words.\\
The following are given passages.\\
\{context\}\\
Answer the question based on the given passages. Only give me the answer and do not output any other words.\\
Question: \{input\}\\
Answer:
\end{tcolorbox}

\begin{tcolorbox}[size=title,opacityfill=0.1,breakable]
\noindent
\textbf{2WikiMultihopQA}: Answer the question based on the given passages. Only give me the answer and do not output any other words.\\
The following are given passages.\\
\{context\}\\
Answer the question based on the given passages. Only give me the answer and do not output any other words.\\
Question: \{input\}\\
Answer:
\end{tcolorbox}

\begin{tcolorbox}[size=title,opacityfill=0.1,breakable]
\noindent
\textbf{MuSiQue}: Answer the question based on the given passages. Only give me the answer and do not output any other words.\\
The following are given passages.\\
\{context\}\\
Answer the question based on the given passages. Only give me the answer and do not output any other words.\\
Question: \{input\}\\
Answer:
\end{tcolorbox}

\begin{tcolorbox}[size=title,opacityfill=0.1,breakable]
\noindent
\textbf{DuReader}: 
\begin{CJK}{UTF8}{gbsn}
请基于给定的文章回答下述问题。\\
文章：\{context\}\\
请基于上述文章回答下面的问题。\\
问题：\{input\}\\
回答：
\end{CJK}
\end{tcolorbox}

\begin{tcolorbox}[size=title,opacityfill=0.1,breakable]
\noindent
\textbf{GovReport}: You are given a report by a government agency. Write a one-page summary of the report.\\
Report:\\
\{context\}\\
Now, write a one-page summary of the report.\\
Summary:
\end{tcolorbox}

\begin{tcolorbox}[size=title,opacityfill=0.1,breakable]
\noindent
\textbf{QMSum}: You are given a meeting transcript and a query containing a question or instruction. Answer the query in one or more sentences.\\
Transcript:\\
\{context\}\\
Now, answer the query based on the above meeting transcript in one or more sentences.\\
Query: \{input\}\\
Answer:
\end{tcolorbox}

\begin{tcolorbox}[size=title,opacityfill=0.1,breakable]
\noindent
\textbf{MultiNews}: You are given several news passages. Write a one-page summary of all news.\\
News:\\
\{context\}\\
Now, write a one-page summary of all the news.\\
Summary:
\end{tcolorbox}

\begin{tcolorbox}[size=title,opacityfill=0.1,breakable]
\noindent
\textbf{VCSUM}: 
\begin{CJK}{UTF8}{gbsn}
下面有一段会议记录，请你阅读后，写一段总结，总结会议的内容。\\
会议记录：\\
\{context\}\\
会议总结：
\end{CJK}
\end{tcolorbox}

\begin{tcolorbox}[size=title,opacityfill=0.1,breakable]
\noindent
\textbf{TREC}: Please determine the type of the question below. Here are some examples of questions.\\
\{context\}\\
\{input\}
\end{tcolorbox}

\begin{tcolorbox}[size=title,opacityfill=0.1,breakable]
\noindent
\textbf{TriviaQA}: Answer the question based on the given passage. Only give me the answer and do not output any other words. The following are some examples.\\
\{context\}\\
\{input\}
\end{tcolorbox}

\begin{tcolorbox}[size=title,opacityfill=0.1,breakable]
\noindent
\textbf{SAMSum}: Summarize the dialogue into a few short sentences. The following are some examples.\\
\{context\}\\
\{input\}
\end{tcolorbox}

\begin{tcolorbox}[size=title,opacityfill=0.1,breakable]
\noindent
\textbf{LSHT}: 
\begin{CJK}{UTF8}{gbsn}
请判断给定新闻的类别，下面是一些例子。\\
\{context\}\\
\{input\}
\end{CJK}
\end{tcolorbox}

\begin{tcolorbox}[size=title,opacityfill=0.1,breakable]
\noindent
\textbf{PassageCount}: There are some paragraphs below sourced from Wikipedia. Some of them may be duplicates. Please carefully read these paragraphs and determine how many unique paragraphs there are after removing duplicates. In other words, how many non-repeating paragraphs are there in total?\\
\{context\}\\
Please enter the final count of unique paragraphs after removing duplicates. The output format should only contain the number, such as 1, 2, 3, and so on.\\
The final answer is: 
\end{tcolorbox}

\begin{tcolorbox}[size=title,opacityfill=0.1,breakable]
\noindent
\textbf{PassageRetrieval-en}: Here are 30 paragraphs from Wikipedia, along with an abstract. Please determine which paragraph the abstract is from.\\
\{context\}\\
The following is an abstract.\\
\{input\}\\
Please enter the number of the paragraph that the abstract is from. The answer format must be like ``Paragraph 1'', ``Paragraph 2'', etc.\\
The answer is:
\end{tcolorbox}

\begin{tcolorbox}[size=title,opacityfill=0.1,breakable]
\noindent
\textbf{PassageRetrieval-zh}: 
\begin{CJK}{UTF8}{gbsn}
以下是若干段落文字，以及其中一个段落的摘要。请确定给定的摘要出自哪一段。\\
\{context\}\\
下面是一个摘要\\
\{input\}\\
请输入摘要所属段落的编号。答案格式必须是``段落1''，``段落2''等格式\\
答案是：
\end{CJK}
\end{tcolorbox}

\begin{tcolorbox}[size=title,opacityfill=0.1,breakable]
\noindent
\textbf{LCC}: Please complete the code given below. \\
\{context\}Next line of code:
\end{tcolorbox}

\begin{tcolorbox}[size=title,opacityfill=0.1,breakable]
\noindent
\textbf{RepoBench-P}: Please complete the code given below.\\
\{context\}\{input\}Next line of code:
\end{tcolorbox}

\xhdr{Maximum Output Length}
We set a maximum output length on each dataset during evaluation to prevent the models from non-stop generation.
\begin{table}[htbp]
\centering
\resizebox{\textwidth}{!}{
\begin{tabular}{ccccccccccccccccccccc}
\toprule
1-1 & 1-2 & 1-3 & 1-4 & 2-1 & 2-2 & 2-3 & 2-4 & 3-1 & 3-2 & 3-3 & 3-4 & 4-1 & 4-2 & 4-3 & 4-4 & 5-1 & 5-2 & 5-3 & 6-1 & 6-2 \\
\midrule
128 & 128 & 64 & 64 & 32 & 32 & 32 & 128 & 512 & 512 & 512 & 512 & 64 & 32 & 128 & 64 & 32 & 32 & 32 & 64 & 64 \\
\bottomrule
\end{tabular}
}
\end{table}

\section{Radar Plot and Analysis}
\begin{figure}[htbp]
    \centering
    \includegraphics[width=\textwidth]{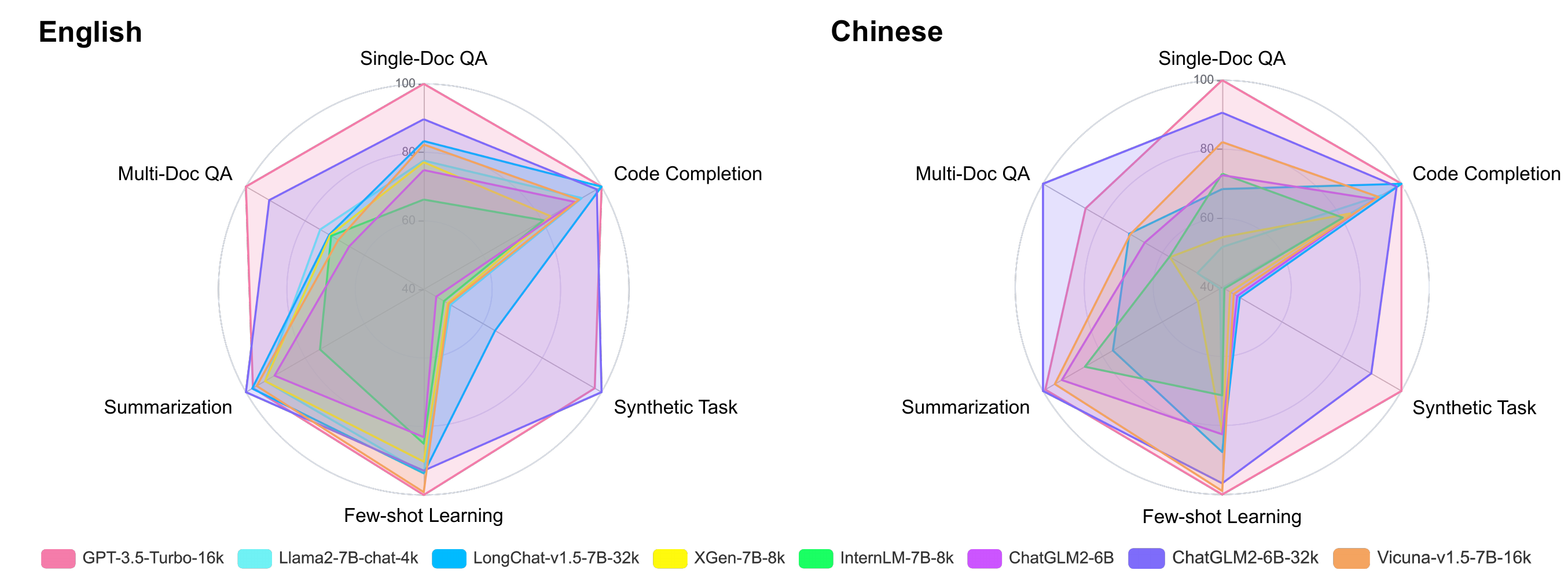}
    \caption{Average scores on 6 major tasks, on English and Chinese datasets, respectively.}
    \label{fig:radar}
\end{figure}
Among the 6 major tasks, summarization and code completion tend not to be sufficiently discerning. This may be due to the fact that the similarity-based metrics (ROUGE-L, Edit Sim) on these tasks are not sensitive enough to well distinguish between the strong and weak models.
Meanwhile, we find that synthetic tasks tend to offer a higher level of discernment, where models either achieve a high score or display a near-zero performance.
These findings lead us to believe that it may not be a good idea to simply obtain an average over all tasks as the sign of models' long context capability, as used in previous benchmark~\citep{shaham2023zeroscrolls} --- since the performance on the more discerning tasks, such as the synthetic tasks in our benchmark, may dominate the final rank.
This necessitates an evaluation strategy like we use in LongBench that separately assesses each task category, potentially leading to more meaningful benchmarking results.

\section{Analysis on the Inter-task Correlation on LongBench}
\label{app:corr}
\begin{figure}[htbp]
    \centering
    \includegraphics[width=0.7\textwidth]{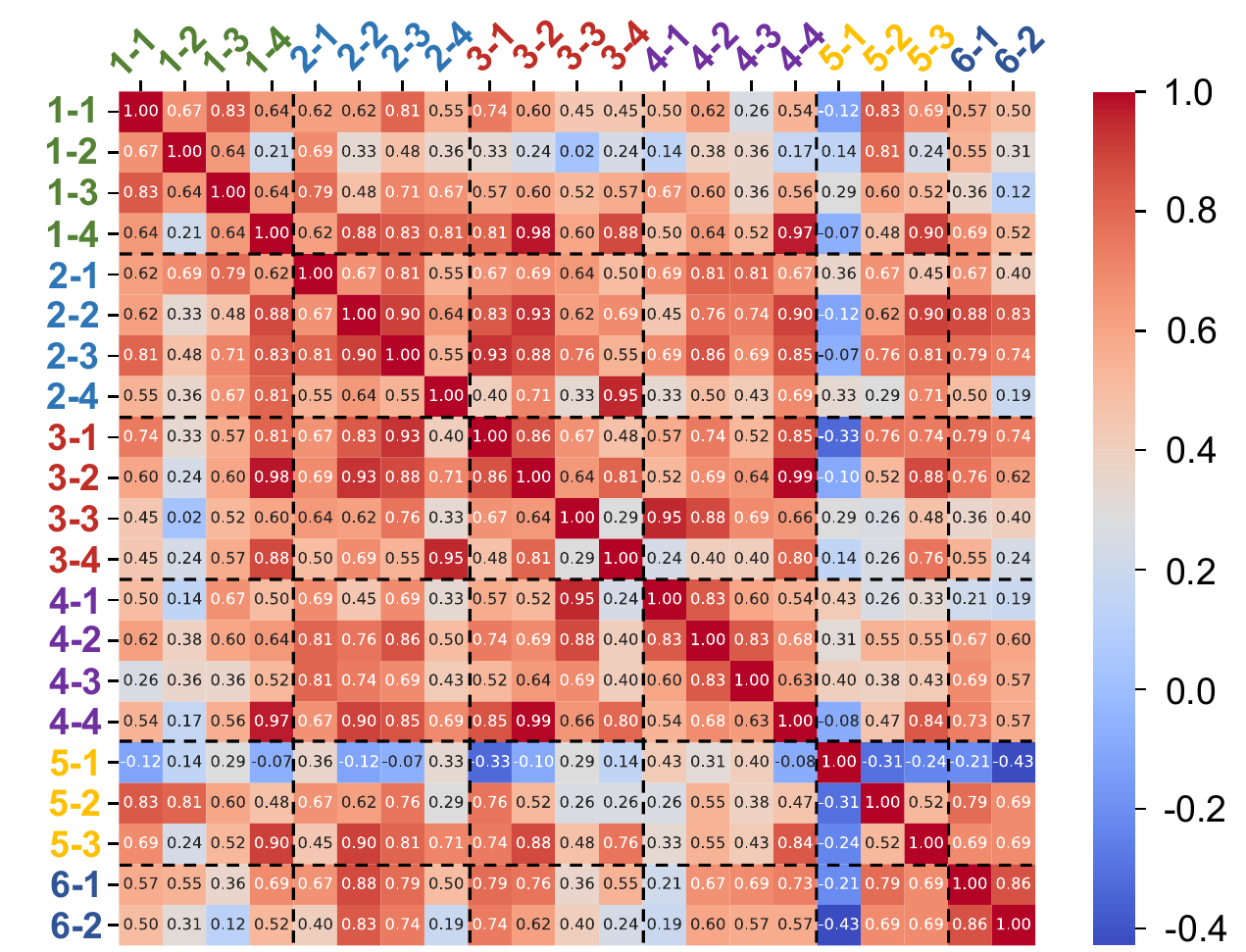}
    \caption{Spearman correlation between each pair of task in LongBench.}
    \label{fig:heatmap}
\end{figure}

We analyze the multi-task property of LongBench by the inter-task correlation among and across each category of tasks, as shown in Figure~\ref{fig:heatmap}.
We observe that most tasks within the same task category have a high correlation, except for PassageCount (5-1), which exhibits low correlation with almost all tasks since models perform poorly (almost random) on this challenging task.
Meanwhile, we notice that the correlations between Qasper (1-2), RepoBench-P (6-2) and the other tasks are also lower, which implies that these tasks potentially require a different attention pattern than the other tasks.
Notably, tasks in the same language have a higher correlation with each other, e.g., a high correlation between Chinese tasks (1-4, 2-4, 3-4, 4-4, 5-3).
These observations suggest that LongBench provides a more comprehensive evaluation result by integrating various types of tasks and languages.

\section{Full results on LongBench-E}
We show the full results on LongBench-E in Table~\ref{tb:exp-e}.

\begin{table}[htbp]
\centering
\resizebox{\textwidth}{!}{
\begin{tabular}{lc|c|cc|cc|cc|ccc|cc|cc}
\toprule
\multirow{2}{*}{\textbf{Model}} & \multirow{2}{*}{\textbf{Length}} & \multirow{2}{*}{\textbf{Avg}} & \multicolumn{2}{c|}{\textbf{S-Doc QA}} & \multicolumn{2}{c|}{\textbf{M-Doc QA}} & \multicolumn{2}{c|}{\textbf{Summ}} & \multicolumn{3}{c|}{\textbf{Few-shot}} & \multicolumn{2}{c|}{\textbf{Synthetic}} & \multicolumn{2}{c}{\textbf{Code}} \\
\cmidrule(lr){4-5} \cmidrule(lr){6-7} \cmidrule(lr){8-9} \cmidrule(lr){10-12} \cmidrule(lr){13-14} \cmidrule(lr){15-16}
& & & \textbf{1-2} & \textbf{1-3} & \textbf{2-1} & \textbf{2-2} & \textbf{3-1} & \textbf{3-3} & \textbf{4-1} & \textbf{4-2} & \textbf{4-3} & \textbf{5-1} & \textbf{5-2} & \textbf{6-1} & \textbf{6-2} \\
\midrule
\multirow{3}{*}{GPT-3.5-Turbo-16k} & 0-4k & 51.5 & 45.8 & 57.4 & 64.6 & 49.8 & 31.3 & 26.9 & 57.7 & 88.1 & 38.1 & 9.8 & 99.0 & 58.8 & 52.0 \\
& 4-8k & 47.4 & 41.1 & 43.0 & 53.0 & 45.1 & 29.6 & 23.4 & 71.7 & 91.6 & 37.1 & 9.5 & 90.7 & 52.2 & 46.9 \\
& 8k+ & 42.4 & 27.9 & 61.8 & 50.9 & 23.6 & 28.4 & 22.6 & 75.3 & 87.4 & 40.6 & 1.1 & 66.7 & 47.8 & 42.4 \\
\midrule
\multirow{3}{*}{Llama2-7B-chat-4k} & 0-4k & 35.9 & 20.9 & 43.5 & 36.8 & 33.3 & 31.7 & 27.1 & 52.0 & 81.9 & 40.6 & 8.3 & 17.0 & 57.5 & 38.8 \\
& 4-8k & 30.5 & 18.0 & 31.5 & 29.2 & 22.5 & 27.8 & 22.9 & 58.0 & 80.4 & 37.0 & 1.9 & 4.0 & 49.2 & 41.8 \\
& 8k+ & 29.6 & 21.1 & 31.1 & 24.4 & 21.5 & 25.6 & 22.0 & 58.0 & 83.4 & 42.1 & 2.8 & 9.0 & 35.6 & 40.4 \\
\midrule
\multirow{3}{*}{LongChat-v1.5-7B-32k} & 0-4k & 36.9 & 28.4 & 44.1 & 30.8 & 26.0 & 34.0 & 27.1 & 50.0 & 81.0 & 38.6 & 0.0 & 35.0 & 50.8 & 54.0 \\
& 4-8k & 35.3 & 27.5 & 37.5 & 34.6 & 18.8 & 30.7 & 23.1 & 65.0 & 81.5 & 31.7 & 0.1 & 22.0 & 60.7 & 50.3 \\
& 8k+ & 34.5 & 14.0 & 48.6 & 25.2 & 19.1 & 28.4 & 22.3 & 61.0 & 86.6 & 32.2 & 0.0 & 25.0 & 60.8 & 50.4 \\
\midrule
\multirow{3}{*}{XGen-7B-8k} & 0-4k & 32.6 & 19.4 & 49.9 & 34.0 & 21.9 & 31.0 & 27.7 & 59.0 & 83.7 & 25.0 & 8.0 & 7.8 & 37.1 & 42.4 \\
& 4-8k & 27.5 & 17.9 & 27.5 & 23.5 & 19.4 & 28.0 & 21.9 & 70.0 & 67.9 & 25.1 & 4.1 & 8.0 & 36.3 & 35.1 \\
& 8k+ & 27.4 & 16.7 & 29.6 & 26.2 & 13.6 & 26.5 & 21.0 & 68.0 & 81.0 & 25.6 & 1.0 & 8.0 & 30.4 & 38.8 \\
\midrule
\multirow{3}{*}{InternLM-7B-8k} & 0-4k & 30.4 & 19.7 & 32.0 & 43.3 & 24.4 & 18.0 & 21.3 & 50.0 & 80.0 & 21.2 & 8.0 & 18.0 & 47.4 & 32.3 \\
& 4-8k & 23.0 & 13.7 & 16.5 & 17.5 & 28.6 & 9.4 & 17.4 & 46.0 & 77.5 & 21.4 & 7.7 & 7.0 & 36.0 & 25.4 \\
& 8k+ & 23.2 & 26.2 & 16.0 & 24.9 & 15.0 & 6.6 & 15.9 & 36.0 & 80.5 & 20.0 & 4.5 & 10.0 & 39.1 & 28.8 \\
\midrule
\multirow{3}{*}{ChatGLM2-6B} & 0-4k & 33.1 & 19.6 & 45.5 & 27.8 & 31.3 & 29.6 & 25.6 & 36.0 & 76.9 & 32.8 & 6.5 & 22.2 & 51.3 & 41.2 \\
& 4-8k & 28.5 & 21.1 & 28.0 & 19.2 & 24.6 & 23.4 & 21.9 & 47.0 & 72.5 & 29.0 & 6.0 & 8.0 & 49.5 & 40.9 \\
& 8k+ & 25.5 & 16.0 & 19.4 & 21.7 & 15.8 & 20.1 & 20.4 & 46.0 & 69.9 & 28.2 & 2.3 & 5.0 & 49.0 & 40.5 \\
\midrule
\multirow{3}{*}{ChatGLM2-6B-32k} & 0-4k & 44.2 & 33.9 & 45.0 & 47.5 & 39.9 & 34.9 & 27.1 & 56.0 & 77.0 & 33.2 & 3.0 & 85.0 & 55.1 & 48.3 \\
& 4-8k & 43.4 & 33.4 & 44.8 & 45.2 & 38.0 & 33.2 & 22.0 & 68.0 & 74.7 & 32.1 & 4.0 & 79.0 & 58.7 & 45.4 \\
& 8k+ & 43.1 & 23.4 & 57.4 & 42.2 & 26.4 & 31.5 & 21.3 & 71.0 & 81.8 & 33.6 & 5.0 & 81.0 & 55.4 & 49.3 \\
\midrule
\multirow{3}{*}{Vicuna-v1.5-7B-16k} & 0-4k & 37.3 & 29.2 & 46.4 & 38.2 & 30.8 & 34.1 & 28.0 & 56.0 & 84.2 & 39.7 & 7.0 & 18.0 & 56.1 & 40.2 \\
& 4-8k & 32.3 & 20.1 & 32.9 & 23.9 & 17.4 & 30.4 & 23.7 & 73.0 & 85.1 & 37.3 & 3.0 & 7.0 & 59.5 & 39.5 \\
& 8k+ & 29.6 & 21.8 & 28.1 & 19.7 & 12.3 & 24.4 & 21.5 & 68.0 & 89.9 & 39.2 & 1.0 & 7.0 & 46.5 & 41.4 \\
\bottomrule
\end{tabular}
}
\caption{Results (\%) on LongBench-E.}
\label{tb:exp-e}
\end{table}


\end{document}